\documentclass[lettersize,journal]{IEEEtran}
\usepackage{amsmath,amsfonts}
\usepackage{algorithmic}
\usepackage{algorithm}
\usepackage{array}
\usepackage[caption=false,font=normalsize,labelfont=sf,textfont=sf]{subfig}
\usepackage{textcomp}
\usepackage{stfloats}
\usepackage{url}
\usepackage{verbatim}
\usepackage{graphicx}
\usepackage{cite}
\usepackage{multirow}
\usepackage{float}
\usepackage{amsmath,amsthm}
\newtheorem{Definition}{Definition}
\newtheorem{Theorem}{Theorem}

\newtheorem{Lemma}{Lemma}
\newtheorem{Proposition}{Proposition}
\usepackage{booktabs} 
\usepackage{color}
\usepackage{makecell}
\begin{document}
\title{CCMamba: Topologically-Informed Selective State-Space Networks on Combinatorial Complexes for Higher-Order Graph Learning }


\author{
Jiawen Chen, Qi Shao, Mingtong Zhou, Duxin Chen, Wenwu Yu,~\IEEEmembership{Senior Member, IEEE,} 
\thanks{This research was supported by the National Science and Technology Major Project of the Ministry of Science and Technology of China (Grant No.2024ZD0608104), the National Natural Science Foundation of China (Grants No.62233004, 62273090, and 62073076), the Zhishan Youth Scholar
Program, the Jiangsu Provincial Scientific Research Center of Applied Mathematics (Grant No. BK20233002), the Natural Science Foundation of Jiangsu Province of China (Grants No. BK20253018, and BK20253020), the Open Research Project of the State Key Laboratory of Industrial Control Technology, China (Grant No. ICT2025B54). (Corresponding authors: Duxin Chen, Wenwu Yu.) }%
\thanks{Jiawen Chen, Qi Shao, Mingtong Zhou, Duxin Chen, Wenwu Yu are with the Jiangsu Key Laboratory of Networked Collective Intelligence, School of Mathematics, Southeast University, Nanjing 210096, China. (e-mail:chenjiawen@seu.edu.cn; shaoqi@seu.edu.cn; 213221863@seu.edu.cn; chendx@seu.edu.cn; wwyu@seu.edu.cn).} 
}

\markboth{Journal of \LaTeX\ Class Files,~Vol.~14, No.~8, August~2021}%
{Shell \MakeLowercase{\textit{et al.}}: A Sample Article Using IEEEtran.cls for IEEE Journals}


\maketitle

\begin{abstract}
Topological deep learning has emerged as a powerful paradigm for modeling higher-order relational structures beyond pairwise interactions that standard graph neural networks fail to capture. While combinatorial complexes (CCs) offer a unified topological foundation for the higher-order graph learning, existing topological deep learning methods rely heavily on local message passing and attention mechanisms. These suffer from quadratic complexity and local neighborhood constraints, limiting their scalability and capacity for rank-aware, long-range dependency modeling. To overcome these challenges, 
we propose Combinatorial Complex Mamba (CCMamba), the first unified Mamba-based neural framework for learning on combinatorial complexes. CCMamba reformulates higher-order message passing as a selective state-space modeling problem by linearizing multi-rank incidence relations into structured, rank-aware sequences. This architecture enables adaptive, directional, and long-range information propagation in linear time bypassing the scalability bottlenecks of self-attention. Theoretically, we further establish that the expressive power of CCMamba is upper-bounded by the 1-dimensional combinatorial complex Weisfeiler-Lehman (1-CCWL) test. Extensive experiments across graph, hypergraph, and simplicial benchmarks demonstrate that CCMamba consistently outperforms existing methods while exhibiting superior scalability and remarkable robustness against over-smoothing in deep architectures.
  
\end{abstract}


\section{Introduction}

\begin{figure}[htbp]
\centering 
\includegraphics[width=1 \linewidth]{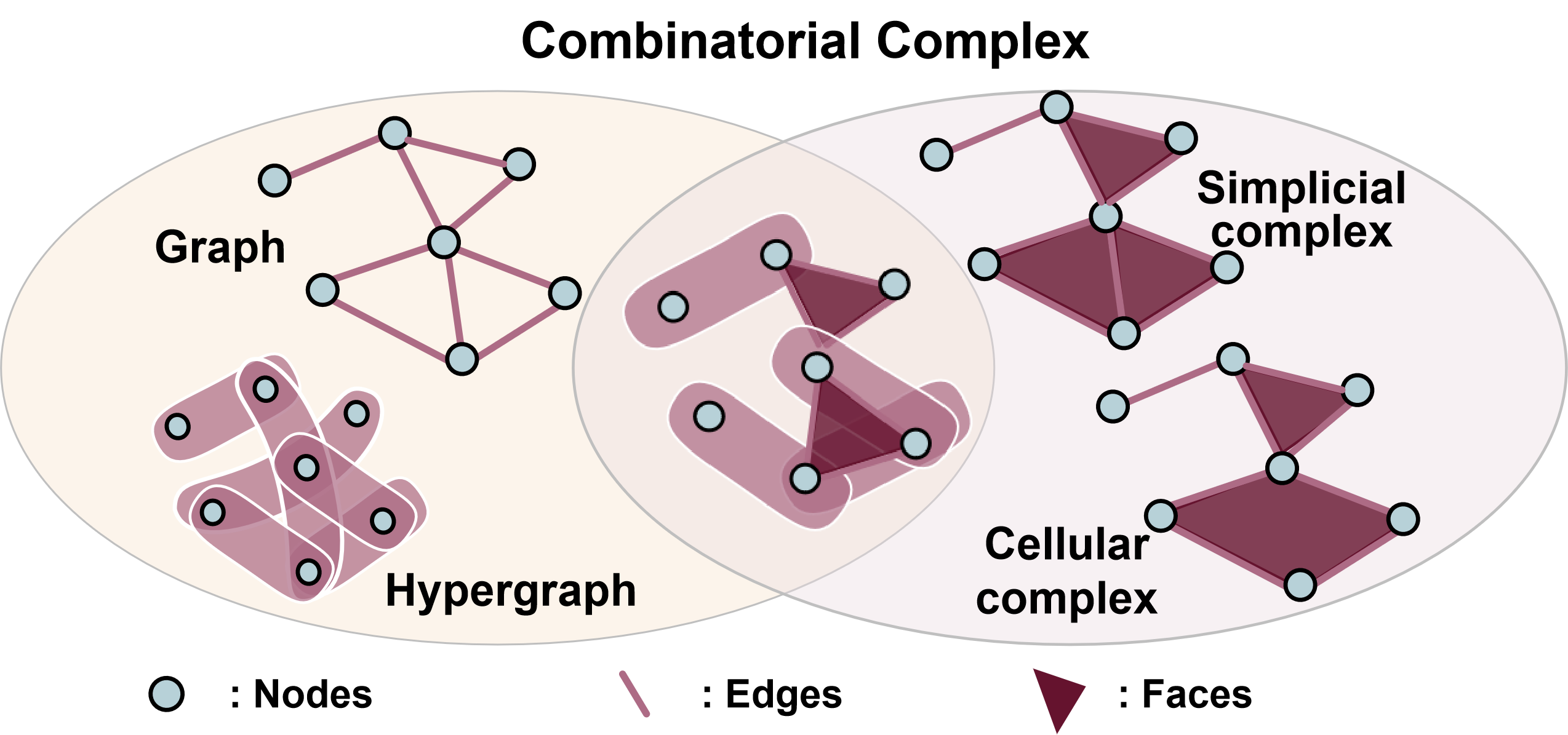}
\caption{Illustration of combinatorial complex}
\label{fig01}
\end{figure}

\IEEEPARstart{T}{opological} relational learning on graphs emerges as a fundamental paradigm for understanding complex systems. 
Many real-world systems, ranging from biomolecular interaction networks~\cite{truong2024weisfeiler} to traffic dynamics~\cite{pham2025topological}, image manifolds~\cite{love2023topological}, and multi-scale geometric structures~\cite{besta2025demystifying,liu2025graph}, exhibit relational patterns that extend far beyond pairwise interactions.
While graphs offer a flexible representation for modeling 1-dimensional relations~\cite{10826583}, 
they are intrinsically limited in expressing multi-way, hierarchical, and boundary-aware dependencies that arise in higher-order domains such as hypergraphs~\cite{8391739, su2025hypergraph}, simplicial complexes~\cite{wu2023simplicial}, and cellular complexes. 
This limitation motivates the proposal of combinatorial complexes (CCs) in Figure~\ref{fig01}, which bridges the gap between cell complexes and hypergraphs~\cite{hajij2023combinatorial}, provides a principled topological framework to capture higher-order interactions. CCs organize entities into ranked cells (nodes, edges, faces, etc.) connected via boundary and co-boundary operators~\cite{hajij2022topological,hajij2023combinatorial,besta2025demystifying}, thereby offering a unified structure that can generalize graphs, hypergraphs, and higher-order complexes.
This framework has become frontier to topological deep learning (TDL).

To exploit these structures, a rich family of neural architectures have been proposed.
Hypergraph neural networks(HGNNs) employ node–hyperedge message passing and have achieved strong performance on node classification tasks~\cite{cai2022hypergraph}, typically through two-stage aggregation schemes~\cite{feng2019hypergraph,ding2020more,ding2023hyperformer}.
However, HGNNs primarily model node–edge interactions and consider only low-dimensional, local message passing, leaving higher-order dependencies (e.g. edge–face, multi-rank interactions) unmodeled.
Unified approaches like UniGNN~\cite{huang2021unignn}, CCNNs~\cite{hajij2022topological,hajij2023combinatorial}, and the topological message passing framework of TopoTune~\cite{papillon2023architectures} attempt to generalize beyond hypergraphs.
These methods introduce higher-order message passing, but most remain iterative and inherently local~\cite{naitzat2020topology,wen2024tensor,hajij2025copresheaf}, limiting their ability to capture long-range, directional and fully structured dependencies.
These advances demonstrate the value of explicitly modeling structural geometry, yet most of them rely on low-order or graph-based formulations. This limitation highlights the need to move beyond simple neighborhood aggregations to exploit the geometry learning of higher-order structures.

Parallel to these developments, a series of recent works has explored new perspectives on topological deep learning and the integration of topological priors into neural architectures. 
Prior works like MPSNs~\cite{bodnar2023topological} and HOMP~\cite{eitantopological} , which extend cellular complexes via algebraic topology, formalize message passing in terms of node-edge interactions.
Beyond hypergraphs, architectures such as simplicial neural networks ~\cite{roddenberry2021principled,lecha2025higher}, SCNNs~\cite{wu2023simplicial}, cellular neural networks~\cite{bodnar2021weisfeiler}, and TopoTune~\cite{battiloro2024latent} extend message passing to simplicial and cellular complexes, leveraging algebraic-topological operators to capture higher-order interactions.
These approaches are tied to specific complex types and overlook asymmetric or directional flows across ranks~\cite{hajij2022topological,hajij2023combinatorial} . 
Recent attempts extend attention and transformer-style architectures~\cite{papamarkou2024position,papillontopotune,bick2024transformers} on hypergraph, simplicial complex. 
However, these attention-based methods suffer from the quadratic complexity of self-attention that results in them difficult to scale to large-scale graphs. Furthermore, existing methods fails to preserve the rank-aware information on  combinatorial complexes.

Selective State-Space models (SSMs)~\cite{daotransformers,yuan2025dg,nishikawastate,lahoti2026mamba} have recently emerged as an alternative to Transformer architectures, which address the transformer's high computation while maintaining the ability to model long sequences. Building on the breakthroughs of selective SSMs, particularly the Mamba-2, we investigate their integration into topological deep learning. 
Unlike the linear or temporal sequences for which SSMs were originally designed, CCs possess a structured, multi-rank topology where information flows across disparate dimensions (e.g., nodes, edges, and faces). Simply substituting attention mechanisms with Mamba layers~\cite{wang2024graph} fails to model rank-aware message propagation, as it ignores the complex boundary and coboundary relations inherent in higher-order structures. Consequently, there is a critical need to reformulate the state-space paradigm: we must move beyond local, isotropic neighborhood aggregations toward a unified, topologically-informed framework that can capture multi-rank structural dependencies through a global, sequence-aware perspective.

To address these limitations, we propose \textbf{Combinatorial Complex Mamba (CCMamba)}, the first unified Mamba-based neural architecture designed for the higher-order graph learning on combinatorial complexes. 
CCMamba introduces a bidirectional, rank-aware state-space module that linearizes the higher-order incidence relations of CCs into structured sequences, which are then processed via selective state-space models.
By treating topological neighborhoods as structured sequences, CCMamba transcends the locality constraints of traditional topological deep learning approaches, enabling directional and global context-aware propagation.
This design facilitates efficient message passing across nodes, edges, faces, and higher-dimensional cells within a single framework. Here, CCMamba serves as a unified backbone that generalizes to graphs, hypergraphs, simplicial complexes, and cellular complexes by instantiating the corresponding incidence matrices.
Our main contributions are summarized as follows:

\begin{itemize}
\item We propose CCMamba, the first Mamba-based neural networks on combinatorial complexes. It provides a unified a higher-order message passing framework capable of graphs, hypergraphs, simplicial and cellular complexes.

\item To address quadratic complexity of self-attention mechanism, CCMamba introduces a rank-structured selective state-space, which linearizes neighborhood sequences and models long-range propagation.

\item We establish a theoretical analysis demonstrating that the expressivity of CCs-based Mamba message passing is bounded above by the 1-dimensional combinatorial complex Weisfeiler–Leman (1-CCWL) test.

\item We conduct extensive experiments on graph, hypergraph, and simplicial benchmarks, where CCMamba achieves superior results on node and graph classification tasks.
\end{itemize}

\section{RELATED WORK}

In this section, we review the related work of higher-order graph neural networks and graph selective state space models.

\subsection{Higher-order Graph Neural Networks.}
To capture higher-order interactions beyond pairwise graphs, higher-order models have been developed. Hypergraph neural networks enable node–hyperedge message passing~\cite{feng2019hypergraph,ding2020more,cai2022hypergraph},but are limited to node–hyperedge relations. Simplicial and cellular neural networks leverage boundary and co-boundary operators to model higher-order interactions ~\cite{roddenberry2021principled,lecha2025higher,wu2023simplicial,bodnar2021weisfeiler,bodnar2023topological}.
Combinatorial complexes ~\cite{hajij2022topological,hajij2023combinatorial} unify these structures into a single hierarchical framework, aiming for graph representation~\cite{papillon2023architectures,papillontopotune}. However, most studies often oversimplify the topologies to hypergraphs or simplicial complexes, limiting their ability to the long-range dependencies. Existing higher-order neural networks are typically designed for specific structures, including hypergraph neural networks~\cite{huang2021unignn,gao2020hypergraph}, simplicial neural networks, and cell complex networks~\cite{tahademystifying, huang2024higher, battiloro2024latent}.
These models rely on incompatible neighborhood definitions and message-passing mechanisms, reflecting different ways of generalizing graphs through hierarchical refinements~\cite{millan2025topology}. To address this, TopoTune~\cite{papillontopotune} studies combinatorial complex neural networks, but simplifies as three neighborhood types and disrupts the higher-order information propagation.
Furthermore, other models ~\cite{hajij2023combinatorial,papillon2023architectures,besta2025demystifying} fail to faithfully preserve the topological neighborhood of combinatorial complexes and limit their ability to learn complex structures and geometries.

Previous neural network architectures for combinatorial complexes have shown the potential to learn rich, higher-order topological structures. However, the broader field of topological dependency learning (TDL) remains fragmented. Several frameworks leverage CCs to design flexible neural networks, including TopoTune~\cite{papillontopotune}, SMCN~\cite{eitantopological}, topological neural networks~\cite{battiloro2024n,eitan2025topological} and modular toolkits for learning on generalized topological domains~\cite{telyatnikov2024topobenchmarkx}. However, the convolutional layers used in these neural networks suffer from oversmoothing issues and quadratic complexity in the attention mechanism.
This highlights a critical need for a principled, expressive, and efficient framework that preserves the multi-rank neighborhoods inherent to combinatorial complexes.

\subsection{Graph Selective State-space Models}

Selective state-space models (SSMs), particularly the Mamba architecture, have emerged as a compelling alternative to transformer for sequence modeling~\cite{liu2025vision, daotransformers,yuan2025dg}. Mamba’s core strength lies in its ability to achieve linear-time inference through input-dependent state transitions while retaining strong long-range modeling capacity. Motivated by these advantages, several works have begun integrating Mamba into the graph domain. For example, Graph-Mamba 
~\cite{wang2024graph} firstly introduces mamba on graph by integrating a Mamba block with the input-dependent node selection mechanism. 
Some attempts
explore Mamba-based architectures for structured data, and adapt SSMs to graph sequences 
~\cite{He1024963ijcai,behrouz2024graph} and temporal graph signals~\cite{li2024state,shin2026graph}. 
These models display improved scalability and long-range dependency modeling but relying on graph neural networks. However,
current graph Mamba methods are confined to pairwise graph structures and overlook higher-order topological relational learning. 
This paper extends the selective state-space models to combinatorial complexes, which fills the gap in a unified framework for topological information passing and message passing. The proposed framework and theoretical results developed in this paper address the open problems and challenges highlighted in the position literature ~\cite{papamarkou2024position,pmlr-v235-papamarkou24a}, including the generalization of topological message passing and the issue of structural invariance in TDL models.

\section{Preliminaries}

In this section, we introduce the preliminaries of background of combinatorial complexes and state space models.


\subsection{Combinatorial Complexes}
A combinatorial complex generalizes graphs, hypergraphs, simplicial complexes, and cellular complexes into a unified algebraic-topological structure. Formally, we define as follow:

\begin{Definition}[\textbf{Combinatorial Complex}]
A combinatorial complex is a triple $(\mathcal{S}, {C}, \mathrm{rk})$ where

(1) $\mathcal{S}$ is a finite set of vertices,

(2) $ {C} \subseteq 2^{\mathcal{S}}$ is a collection of {cells},

(3) $\mathrm{rk}: {C} \to {Z}_{\ge 0}$ is an order-preserving rank function such that
 $\sigma \subseteq \tau$, then $ \mathrm{rk}(\sigma) \le \mathrm{rk}(\tau)$
\end{Definition}

The elements of $\mathcal{C}$ are called cells (i.e., group of nodes, edges). The rank of a cell $\sigma \in \mathcal{C}$ is $k: \mathrm{rk}(\sigma)$, and we denote it as a $k-$cell, such as 0-cells (nodes), 1-cells (edges), and 2-cells (faces), which simplifies notation for $(\mathcal{S}, {C}, \mathrm{rk})$ , and its dimension are the maximal rank: $\text{dim}(\mathcal{C}):=\text{max}_{\sigma \in \mathcal{C}} \mathrm{rk}(\sigma)$. 


To enable the representation of higher-order structures construction, 
we define a lifting function $f: G\to \mathcal{CC}$ to introduce this lifting operator~
\cite{telyatnikov2024topobenchmarkx, franco2026differentiable} process graph data.
 
\begin{Definition}[Lifting Operation]
Let $\mathcal{G}$ be the class of graphs and let
$\mathcal{H}, \mathcal{S}, \mathcal{C}$ denote the classes of hypergraphs,
simplicial complexes, and cellular complexes, respectively. 
A {lifting operation} is a function
$
f:\mathcal{G}\to \mathcal{X},
$
which lifts graph $G\in\mathcal{G}$ to higher-order graph 
$f(G)\in\mathcal{X}=\{\mathcal{H},\mathcal{S},\mathcal{C}\}$ obtained from $G$ by a prescribed construction.
\end{Definition}

For instance, algorithm \ref{alg:lifting} provides the details of a lifting operator from graph to combinatorial complex.

\begin{algorithm}[htbp]
\caption{Lifting from Graph to Combinatorial Complex }\label{alg:lifting}
\begin{algorithmic}[1]
 
\STATE \textbf{Input:} Graph $G = (V, E)$,  features $\mathbf{X}$, Maximum rank $K$, Neighborhood types $\mathcal{N}$ and Maximum cell length $L$
\STATE \textbf{Output:} Combinatorial complex $\mathcal{CC}$, Incidence matrix $\mathcal{M}$ \\
\STATE Initialize $\mathcal{CC}$$ \leftarrow$$ \emptyset$ and incidence matrices $\mathcal{M} $$\leftarrow $$\emptyset$ 
\STATE {\textbf{Phase 1}: Structural Feature Mapping}

\STATE \hspace{0.5cm} \textbf{for} each node $v \in V$ \textbf{do}
\STATE \hspace{1.0cm} Extract node features $\mathbf{X}_1$ from $\mathbf{X}$ 
\STATE \hspace{1.0cm} Add $0$-cell to $\mathcal{CC}$ with node features  \\
\STATE \hspace{0.5cm} \textbf{end for}
 
\STATE \hspace{0.5cm} \textbf{for} each edge $e=(u, v) \in E$ \textbf{do}
\STATE \hspace{1.0cm} Add $1$-cell to $\mathcal{CC}$ from $E$ with edge features 
\STATE \hspace{0.5cm} \textbf{end for}
\STATE {\textbf{Phase 2}: Extract Higher-Order Structure and Lifting}
\STATE \hspace{0.5cm} \textbf{for} $k = 2$ \textbf{to} $K$ \textbf{do}
\STATE \hspace{1.0cm} $\mathcal{S}_k \leftarrow$ Identify $k$-cell with $L$ (cliques, cycles) \\
\STATE \hspace{1.0cm} Add cells in $\mathcal{S}_k$ to $\mathcal{C}$ as $k$-th rank cells
\STATE \hspace{0.5cm} \textbf{end for}
 
\STATE {\textbf{Phase 3}: Neighborhood Incidence Construction}

\STATE \hspace{0.5cm} \textbf{for} rank $r = 0$ \textbf{to} $K$ \textbf{do}
\STATE \hspace{1.0cm} $B_{r} \leftarrow \text{Incidence Matrix}(r-1, r) \in \mathbb{R}^{n_{r-1} \times n_r}$
\STATE \hspace{1.0cm} $L_{r} \leftarrow \text{Hodge Laplacian}(r)$
\STATE \hspace{1.0cm} $\mathcal{M} \leftarrow \mathcal{M} \cup \{B_r, L_r\}$ \COMMENT{Store incidence matrix}
\STATE \hspace{0.5cm} \textbf{end for}

\STATE \textbf{return} Combinatorial complex $\mathcal{C}$, Incidence matrix $\mathcal{M}$
\end{algorithmic}
\label{alg:lifting}
\end{algorithm}

\subsection{Combinatorial Complex Neural Networks}
Combinatorial complex neural networks (CCNNs) generalize the classical message-passing frameworks to CCs. CCNNs extend message passing beyond pairwise relationships, which define a three-stage message passing mechanism, divide the geometric structure into multiple cell message passing mechanisms, and perform intra-neighbor aggregation and inter-neighbor fusion. Formally, we define as

\paragraph{Combinatorial Complex Neural Networks.}
Let $\mathcal{N}_{\mathcal{C}}$ be a set of neighborhoods, each defines a specific relational operator on the combinatorial complex $\mathcal{CC}$. 
For each cell $\sigma \in {C}$ with rank $\mathrm{rk}(\sigma)$ and feature representation $h_{\sigma}^{l} \in {R}^{F^{l}}$ at layer $l$, the layer update of the CCNNs can be defined as:
\begin{equation}
\label{ccmpnn}
h_{\sigma}^{l+1}
= \phi\ \Big(
 h_{\sigma}^{l},
 \bigotimes_{\mathcal{N}\in\mathcal{N}_{\mathcal{C}}}
 \bigoplus_{\tau\in\mathcal{N}(\sigma)}
 \psi_{\mathcal{N},\,\mathrm{rk}(\sigma)}(h_{\sigma}^{l}, h_{\tau}^{l})
\Big),
\end{equation}
where $\bigoplus$ is {intra-neighborhood aggregation} over all neighbors $\tau \in \mathcal{N}(\sigma)$;
$\bigotimes$ is inter-neighborhood fusion across all neighborhood $\mathcal{N} \in \mathcal{N}_{\mathcal{C}}$;
$\psi_{\mathcal{N},k}$ is a neighborhood-specific {message function}; $\phi$ is the update function. The rank function guarantees message aggregation within each cells neighborhood of different topological ranks.

The message function $\psi_{\mathcal{N},k}$ is parameterized as:
\begin{equation}
\label{eq:msg}
\psi_{\mathcal{N},k}(h_\sigma^l, h_\tau^l)
= \alpha_{\mathcal{N}}(\sigma,\tau)\,
 W_{\mathcal{N},k}\,[\,h_\sigma^l \Vert h_\tau^l\,],
\end{equation}
where $[h_\sigma^l \Vert h_\tau^l]$ is the concatenation of sender and receiver features, $W_{\mathcal{N},k} \in {R}^{F^{l+1} \times 2F^{l}}$ is a learnable transformation matrix, and $\alpha_{\mathcal{N}}(\sigma,\tau)$ is computed as attention mechanism~\cite{ballester2024attending} and~\cite{papillontopotune,gurugubelli2023sann} between cells information propagation.
One observes that scaled dot-product attention coefficient \(\alpha_{\mathcal{N},\tau}^{(h)}\) of cell features \(h\) with neighborhood type \(\mathcal{N}\) and rank \(k = \mathrm{rk}(\sigma)\) is formulated as 
\begin{equation}
\begin{aligned}
 \alpha_{\mathcal{N},k} &= \text{softmax}
 \left( \frac{ \left(W_{Q,\mathcal{N},k} \tilde{h}_\sigma \right) \left( W_{K,\mathcal{N},k} \tilde{h}_\tau \right) }{\sqrt{d/H}} \right) ,
\end{aligned}
\end{equation}
where \(H\) is the number of attention heads, and \(W_{Q,K,V,\mathcal{N},k}^{(h)} \in {R}^{\frac{d}{H} \times d}\) are the query, key, and value projection matrices associated with neighborhood type \(\mathcal{N}\), rank \(k\), and head \(h\). However, applying attention mechanisms to graphs, especially higher-order graphs, results in a high complexity of $O(n^2)$ when calculating the attention score for all node pairs, failing to apply to large-scale real-world networks.



\subsection{State Space Models}
Although attention mechanism effectively captures graph neighborhood dependencies, its quadratic computational complexity imposes computational and memory bottlenecks on CCs.
Recently, the state-space models~\cite{daotransformers,lahoti2026mamba} have emerged as an alternative to address the transformer's high computation while maintaining the ability to model long sequences . State space models (SSM) are linear time-invariant systems, which maps the sequence $x(t)\in R^{L}$ to the embeddings $y(t)\in R^L$. SSM employs a latent state $h(t)\in R^{N\times L}$, parameter matrices $A\in R^{N\times N}$ and $B\in R^{N\times 1}, C\in R^{1\times N}$, defined as 
\begin{equation}
\begin{aligned}
 h'(t) = A\cdot h(t) + Bx(t),\quad
 y(t) = Ch(t).
\end{aligned}
\end{equation}
Furthermore, a discrete spatial state model is proposed to solve the information transfer in deep learning.
\begin{equation}
\begin{aligned}
 h_t = \bar{A} \cdot h_{t-1} + \bar{B}x_t, \quad
 y(t) = Ch(t),
\end{aligned}
\end{equation}
where $\bar{A}=\text{exp}(\Delta A)$, and $\bar{B}=(\Delta A)^{-1}(\text{exp}(\Delta A -I))\cdot \Delta B$.
The selective state-space mechanism of SSM dynamically and selectively propagates information, enabling its computational complexity $O(n)$ scales linearly with the sequence length.

\section{METHODOLOGY}

In this section, we provide a detailed elaboration of the proposed CCMamba and its components.

\subsection{Overview}
\begin{figure*}[htbp]
 \centering \includegraphics[width=0.95\linewidth]{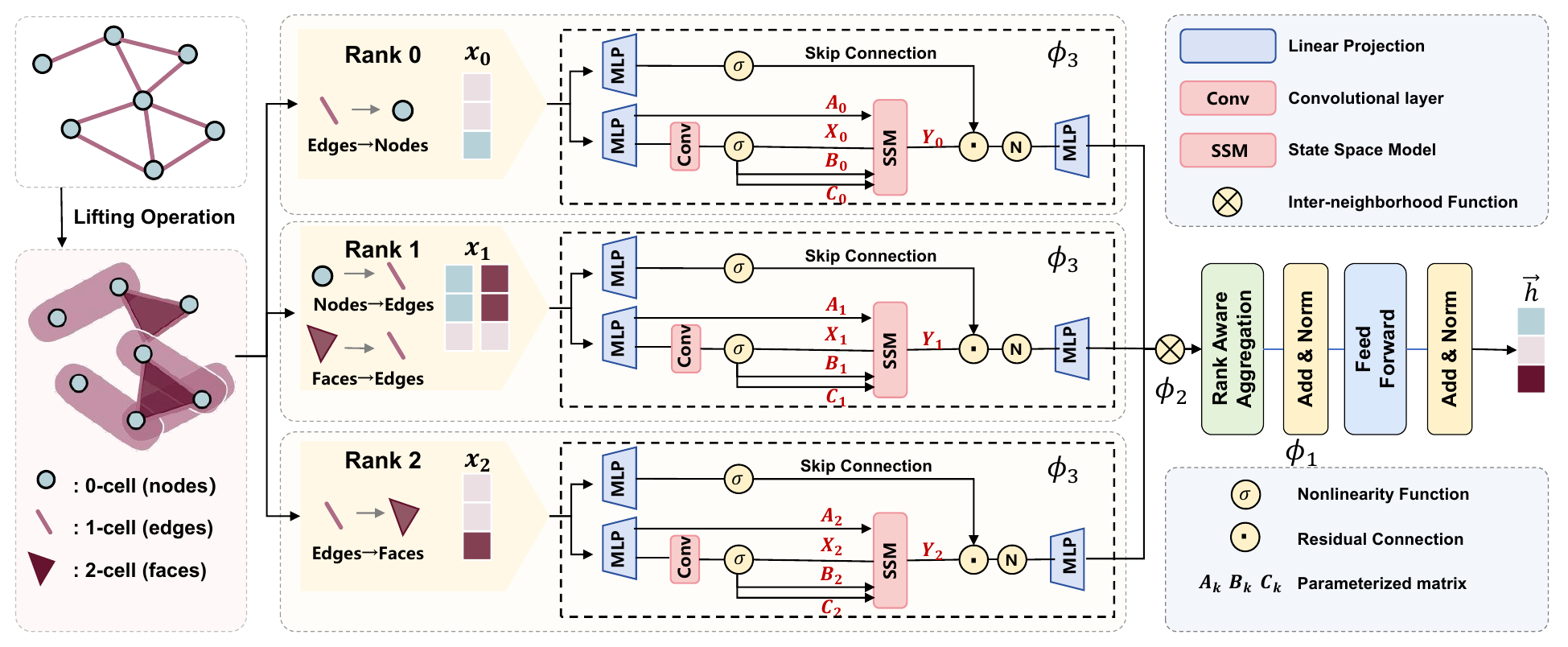}
 \caption{The architecture of the Combinatorial Complex Mamba (CCMamba) framework. CCMamba operates through three hierarchical stages: (1) Topological Construction: For each rank $k \in \{0, 1, 2\}$, incidence-induced neighborhoods are established via boundary and co-boundary operators. (2) Bidirectional Selective Modeling: These topological signals are linearized into structured sequences and processed by rank-specific bidirectional Selective State-Space Models (SSMs) to capture multi-directional dependencies. (3) Inter-rank Fusion: The refined intra-rank features are integrated across adjacent ranks using fusion operators $\phi_1, \phi_2$, and $\phi_3$. The framework incorporates Layer Normalization, Dropout, and Feed-Forward Networks (FFN) to produce the final cell embeddings.}
 \label{fig02}
\end{figure*}

The proposed CCMamba framework is illustrated in Figure~\ref{fig02}, which aims to unify the higher-order graph message passing of combinatorial complexes with the linear-time efficiency of selective state-space models. CCMamba reformulates higher-order relational learning as a suite of rank-aware sequence modeling tasks, bypassing the quadratic complexity and local constraints of traditional attention-based topological models. The framework operates through three primary stages:  
\begin{enumerate}
\item \textbf{Intra-neighborhood Selective SSMs Filter}: This stage leverages incidence-induced relations to construct rank-specific neighborhood representations. By linearizing these topological structures into structured sequences, we employ bidirectional Mamba blocks to perform input-dependent, selective filtering. This mechanism enables the model to capture multi-directional and long-range dependencies within each rank while maintaining a linear computational footprint.

\item \textbf{Inter-neighborhood Rank-level Aggregation}: Following intra-rank refinement, a cross-rank fusion mechanism integrates information from incident cells across different dimensions (e.g., node-to-edge or edge-to-face). By aggregating signals via incident operators, this stage explicitly preserves the hierarchical geometry of the CC and ensures that contextual signals propagate fluidly through the topological ranks hierarchy.

\item \textbf{Deep Representation Refinement}: The final stage utilizes a robust integration pipeline consisting of residual connections, layer normalization, and position-wise Feed-Forward Networks. This architecture facilitates stable gradient flow and enhances the final cell embeddings for downstream classification tasks.

\end{enumerate}

\subsection{Combinatorial Complex State Space Models}

\subsubsection{Neighborhood on Combinatorial Complexes.}
Let $\mathcal{CC}=(\mathcal{S},\mathcal{C},rk)$ denote a 2-dimensional combinatorial complex, where cells of varying dimensions interact through generalized adjacency relations beyond pairwise connections.
we consider incidence-induced, rank-coupled neighborhoods derived from the lower and upper incidence matrices:
$B_1\in{R}^{|V_0|\times|E_1|}$ encodes node–edge incidences and $B_2\in{R}^{|E_1|\times|F_2|}$ represents edge–face incidences. Here, we define four neighborhood types that govern message propagation across dimensions:

1. Node–Edge neighborhoods $\mathcal{N}_{0\rightarrow1}(e)$: the set of nodes incident to edge, capture the impact of 1-cells shared 0-cells. 

2. Edge–Node neighborhoods $\mathcal{N}_{1\rightarrow0}(v)$: the set of edges incident to node $v$, reflecting the 1-cells adjacent to a 0-cell. 

3. Edges–Face neighborhoods $\mathcal{N}_{2\rightarrow1}(e)$: the set of 2-cells that contain a given edge $e$, enabling higher-order interactions among edges via shared faces

4. Face–Edge neighborhoods $\mathcal{N}_{1\rightarrow2}(f)$: the set of edges that bound a face $f$, facilitating information flow from 1-cells to their enclosing 2-cells.

To model these neighborhoods are topologically induced by the rank function of cell. we define neighbor function as

\textbf{Rank-Aware Neighborhood Function}. 
Given a cell $\sigma \in \mathcal{CC}$, let $\mathcal{R}(\sigma)\subseteq\{0,1,2\}$ denote the neighborhood set of three rank types incident to $\sigma$.
Denote by $h^{(\ell)}_\sigma$ the feature of $\sigma$at layer $\ell$,
and its linearly projected feature $\tilde h^{(\ell)}_\sigma=W_k h^{(\ell)}_\sigma$.
we reformulate CCNNs layer with rank functions as
\begin{equation}
\small
h_{\sigma}^{(\ell+1)}
= \phi_1\ \left(
h_{\sigma}^{(\ell)}, \phi_2
\left( \phi_3\left( h_{\sigma}^{(\ell)},\{\tilde h^{(\ell)}_\tau\}_{\tau\in \mathcal{N}_{k\rightarrow r}(\sigma)} \right) \right)_{r\in\mathcal{R}(\sigma)}
\right),
\label{eq:cc-neighborhood-encoding}
\end{equation}
where $\phi_1,\phi_2,\phi_3$ aggregates projected neighbor features, and $\mathcal{N}_{k\rightarrow r}(\sigma) $ is the propagate information from $k$-cell to $r$-cell, and $\phi$ represents a message passing neural network to aggregate information from different neighbors with ranks.

\subsection{Combinatorial Complex Mamba Networks}
While combinatorial complex message-passing networks enable higher-order interaction modeling, their reliance on attention-based neighborhood fusion incurs quadratic complexity and struggles to model long-range dependencies. To address this, we propose Combinatorial Complex Mamba in Figure~\ref{fig02}, a novel framework that integrates selective state-space models into the topological message-passing process.

\paragraph{\textbf{Mamba Block Message Passing Layer}}
Each Mamba block is formulated as a learnable linear-time recurrent filter based on a selective state-space model (SSM), the state transition dynamics are explicitly conditioned on the input features.
For each rank $k$ at layer $\ell$, the SSM computes its representation $h^{(\ell)}_{\text{intra},k}$ via the following parameterization:
\begin{equation}
\label{eq07}
 \begin{aligned}
A_k & = -\exp\!\big(W_{A,k} h^{(\ell)}_k + b_{A,k}\big), 
B_k = W_{B,k} h^{(\ell)}_k, \\
C_k & = W_{C,k} h^{(\ell)}_k,
\Delta_k = \mathrm{Softplus} \big(W_{\Delta,k} h^{(\ell)}_k + b_{\Delta,k}\big), \\
y_k & = \mathrm{SSM}_{\mathrm{selective}}\!\left(A_k, B_k, C_k, \Delta_k, h^{(\ell)}_k\right), \\
z_k & = \sigma \big(W_{z,k} h^{(\ell)}_k\big), h^{(\ell)}_{\text{intra},k} = W_{\mathrm{out},k}\big(z_k \odot y_k\big),
\end{aligned}
\end{equation}
where $A_k$, $B_k$, and $C_k$ respectively parameterize the state transition, input injection, and readout operators of the SSM, while $\Delta_k$ controls the discretization step size and thereby the effective temporal scale of the dynamics. Softplus is the activation function defined as $\text{Softplus(x)} = \ln(1 + e^x)$.
The negative exponential in $A_k$ enforces strictly negative transition coefficients, which guarantees the stability of the continuous-to-discrete state evolution.
The gating vector $z_k$ modulates the SSM output through element-wise multiplication before the final linear projection $W_{\mathrm{out},k}$.
This formulation yields stable propagation dynamics for each rank, with linear-time complexity in the number of cells.

\paragraph{Intra-neighborhood Message Passing via SSM}
Each layer updates the rank-specific features $(h^{(\ell)}_0, h^{(\ell)}_1, h^{(\ell)}_2)$
for nodes, edges, and faces, respectively. 
Let $B_1 \in {R}^{|V_0|\times|E_1|}$ and $B_2 \in {R}^{|E_1|\times|V_2|}$ 
be node–edge and edge–face incidence matrices, respectively. 
Due to topological neighborhoods in combinatorial complexes inherently lack a canonical causal ordering. Linearizing these higher-order structures into 1D sequences forces an arbitrary permutation, potentially leading to information loss and topological bias. To guarantee that information propagation is robust to this linearization, CCMamba integrates a bidirectional state-space module (BiMamba). This ensures that message passing captures structural dependencies from both directions, preserving the geometric fidelity of the incidence relations.
With these operators, the function $\phi_3$ of CCMamba update features at layer $\ell$ can be computed as 
\begin{equation}
\small 
\label{rankssm}
\begin{aligned}
h_{\text{intra},0}^{(\ell+1)} &= \mathrm{MLP}_0^{(\ell)} \left(
\mathrm{BiMamba}_0^{(\ell)} \big(
\mathrm{W}^{(\ell)}_0 [h_0^{(\ell)},\ B_1,h_1^{(\ell)}]
\big)
\right), \\
h_{\text{intra},1}^{(\ell+1)} &= \mathrm{MLP}_1^{(\ell)} \left(
\mathrm{BiMamba}_1^{(\ell)} \big(
\mathrm{W}^{(\ell)}_1[h_1^{(\ell)},\ B_1^{\top} h_0^{(\ell)},\ B_2,h_2^{(\ell)}]
\big)
\right), \\
h_{\text{intra},2}^{(\ell+1)} &= \mathrm{MLP}_2^{(\ell)} \left(
\mathrm{BiMamba}_2^{(\ell)} \big(
\mathrm{W}^{(\ell)}_2 [h_2^{(\ell)},\ B_2^{ \top} h_1^{(\ell)}]
\big)
\right), 
\end{aligned} 
\end{equation}
where $\mathrm{W}^{(\ell)}_k(\cdot)$ and $\mathrm{MLP}^{(\ell)}_k(\cdot)$ are rank-specific linear projections
that align feature dimensions before and after the Mamba blocks. 
Each $\mathrm{BiMamba}_k$ employs a bidirectional filtering mechanism, which computes the hidden states from both the forward process and the reverse process simultaneously, producing direction-aware and rank-specific representations.
This update design implements three coupled message flows:
(1) node–node interactions mediated through shared edges $(B_1 x_1)$,
(2) edge–edge interactions through shared nodes and faces $(B_1^{\top}x_0, B_2 x_2)$, and
(3) face–edge interactions via boundary relations $(B_2^{\top}x_1)$.


\paragraph{Inter-neighborhood Rank-Level Aggregation.}
Following the rank-wise selective state-space propagation in Equation~\ref{rankssm}, 
CCMamba aggregates information across different topological ranks to generate contextually enriched embeddings for each cell. 
Specifically, for a cell $\sigma$ 
, $\mathcal{R}(\sigma)$ denotes the set of ranks adjacent to $\sigma$ via incidence relations (e.g., a edge is adjacent to nodes and faces). Let $\mathbf{h}_{\text{intra},\sigma}^{(\ell)}$ be the representation of $\sigma$ after intra-rank propagation, and $\tilde{\mathbf{H}}_{\text{intra},\mathcal{N}_r(\sigma)}^{(\ell)}$ denote the collection of neighbor representations $h_{\text{intra},r}^{(l)}$ at rank $r$. For $k$-cell, the embedding function $\phi_2$ at layer $\ell+1$ is computed with the summation function, defined as
\begin{equation} 
\label{mamba-rankagg} 
h_{k}^{(\ell+1)} = \phi_1 \Big( \mathbf{h}_{k}^{(\ell)},\; \text{MLP} \Big( \sum_{r \in \mathcal{R}(\sigma)} \big( \tilde{\mathbf{h}}_{\sigma}^{(\ell)},\, \tilde{\mathbf{H}}_{\text{intra},\mathcal{N}_r(\sigma)}^{(\ell)} \big) \Big) \Big),
\end{equation}
where $\phi_1$ is a nonlinear update function (e.g., an MLP layer), and $\mathrm{Mamba}_{r}$ denotes the rank-specific selective state-space module that dynamically filters and integrates cross-rank contextual signals. This construction provides a unified state-space interpretation of higher-order 
message passing while remaining faithful to the combinatorial complex.


After computing rank-wise outputs $h_{\text{intra},0}^{(\ell+1)}, h_{\text{intra},1}^{(\ell+1)}, h_{\text{intra},2}^{(\ell+1)}$, the CCMamba layer refines each representation through residual connections, dropout, and layer normalization:
\begin{equation}
\small
\begin{aligned}
\hat{h}_k^{\,(\ell+1)} &= 
\mathrm{LayerNorm}\ \big(h_{k}^{(\ell+1)} + \mathrm{Dropout}(h_{\text{intra},k}^{(\ell+1)})\big), \\
h_k^{(\ell+1)} &= 
\mathrm{LayerNorm}\ \big(\hat{h}_k^{\,(\ell+1)} + \mathrm{FFN}(\hat{h}_k^{\,(\ell+1)})\big),
\end{aligned}
\end{equation}
where each rank $k \in \{0,1,2\}$ (nodes, edges, faces). This post-processing preserves the incidence-aware structure learned in Equations~\ref{ccmpnn} and~\ref{mamba-rankagg}, while stabilizing training through residual learning. The cell embeddings $\{h_k^{(L)}\}_{k=0,1,2}$ at the final layer $L$ are then pooled to produce representations for downstream tasks such as node classification or graph-level prediction. For an intuitive workflow of CCMamba, we summarize the framework in pseudo-codes in Algorithm \ref{alg:ccmamba_refined}.

\begin{algorithm}[htbp]
\caption{CCMamba Layer-wise Forward Propagation}\label{alg:ccmamba_refined}
\begin{algorithmic}[1] 
\STATE \textbf{Input:} Initial features $h^{(0)} = \{h^{(0)}_0, h^{(0)}_1, h^{(0)}_2\}$; Incidence matrices $B_1 $ and $B_2$; Layers Number $L$; Rank-specific Mamba blocks $\{\text{Mamba}_k\}_{k=0}^2$. \\
\STATE \textbf{Output:} Updated features $h^{(L)} = \{h_0^{(L)}, h_1^{(L)}, h_2^{(L)}\}$
\STATE $h^{(l)} \leftarrow $ $h^{(0)}$ Initialize hidden states across all ranks 
\FOR{ $l = 0$ : $L-1$}
 \STATE \textbf{Step 1}: Intra-neighborhood Features Aggregation  \STATE $h_{\text{intra},k}^{(\ell+1)}$ $ \leftarrow$ $ \text{MLP}_k^{(\ell)}(\text{Mamba}_k^{(\ell)}(W_k^{(\ell)} I_k))$ by applying selective SSM filtering in equations \eqref{eq07}-\eqref{rankssm} 

 \STATE $h_0$ $\leftarrow$ $[h_0^{(l)}, B_1 h_1^{(l)}]$ Aggregate edges signals via $B_1$ 
 \STATE $h_1$ $\leftarrow$ $[h_1^{(l)}, B_1^T h_0^{(l)}, B_2 h_2^{(l)}]$ {Aggregate from edges- nodes signals via $B^T_1$, edges-faces signals via $B_2$}
 \STATE $h_2$ $\leftarrow$ $[h_2^{(l)}, B_2^T h_1^{(l)}]$ {Aggregate faces signals via $B^T_2$}
 
 

 \FORALL{rank $k \in \{0, 1, 2\}$}
 \STATE \textit{// Bidirectional Mamba Message Passing}
  \STATE $\overrightarrow{h}^{(l)}_k$ $ \leftarrow$ $ \text{MLP}_k^{(l)}(\overrightarrow{\text{Mamba}}_k^{(l)}(W_k^{(l)} h^{(l)}_k))$
 \STATE  $ {h}_{\text{intra},k}^{(l+1)}$ $ \leftarrow$ $ \text{MLP}_k^{(l)}(\overleftarrow{\text{Mamba}}_k^{(l)}(W_k^{(l)} \overrightarrow{h}^{(l)}_k)$

 \ENDFOR

 \STATE \textbf{Step 2}: Inter-neighborhood rank-Level Aggregation to generate rank-crossed embeddings in equation \eqref{mamba-rankagg} 

 \FORALL{rank $k \in \{0, 1, 2\}$}
  \STATE $m_k^{(l+1)}$ $ \leftarrow$ $ \text{MLP}(\sum_{r \in \mathcal{R}(\sigma)} \text{Combine}(h_{\text{intra},k}^{(l+1)}, h_{\text{intra},r}^{(l+1)}))$
  \STATE $h_{k}^{(l+1)}$ $ \leftarrow$ $ \phi_1(h_k^{(l)}, m_k^{(l+1)})$ \COMMENT{Non-linear update}
 \ENDFOR

 \STATE $\hat{h}_k^{(\ell+1)} \leftarrow \text{LayerNorm}(h_{k}^{(\ell+1)} + \text{Dropout}(h_{\text{intra},k}^{(\ell+1)}))$
 
 \STATE $h_k^{(\ell+1)} \leftarrow \text{LayerNorm}(\hat{h}_k^{(\ell+1)} + \text{FFN}(\hat{h}_k^{(\ell+1)}))$ 
\ENDFOR

\STATE \textbf{return} $h^{(L)}= \{h_0^{(L)}, h_1^{(L)}, h_2^{(L)}\}$

\end{algorithmic}
\end{algorithm}

\section{How Powerful are CCMamba?}
In this section, we analyze the CCMamba layer through the Weisfeiler–Lehman test with rank-aware cell designed for combinatorial complexes state space models.

\subsection{Generalized Weisfeiler-Leman Algorithm}
The expressive power of message-passing graph neural networks is fundamentally limited by the Weisfeiler–Lehman (WL) hierarchy: standard GNNs are at most as powerful as the 1-WL test~\cite{xupowerful}, while hypergraph GNNs often fail to exceed this bound due to their reliance on symmetric neighborhood aggregation. In contrast, CCs support richer incidence structures that enable more refined topological discrimination. This raises a critical question: How expressive is CCMamba in the higher-order graph learning ?

 \begin{Definition}[(\textbf{Labeled Combinatorial Complex})]
A labeled combinatorial complex $(C,\ell)$ is a $\mathcal{CC}$ equipped with a cell coloring $\ell: \mathcal{C}\to {N}$ . We define $\ell(\sigma)$ is the label coloring of a cell $\sigma\in \mathcal{C}$ on combinatorial complex.
\end{Definition}

\begin{Definition}[\textbf{Combinatorial Complex Weisfeiler Leman (CCWL)}]
Given a 
$\mathcal{CC}(\mathcal{S},\mathcal{C},{rk},\ell^v, \ell^e, \ell^f)$ 
and a function 
$ r: (\mathcal{S},\mathcal{C},{rk},l_i^v, l_i^e, l_i^f) $
$\to$ $(\mathcal{S},\mathcal{C},{rk},l_{i+1}^v, l_{i+1}^e,l_{i+1}^f)$, here $r$ is colorings relabel function. 
we denote by $\mathcal{CC}_i$ the relabeled combinatorial complex after $i$ iterations. 
Then the Combinatorial Complex WL sequence is defined as
\[
\{\mathcal{CC}^{(0)}, \dots, \mathcal{CC}^{(h)}\} 
= \{(\mathcal{S},\mathcal{C},{rk},\ell^{(i)}_v, \ell^{(i)}_e, \ell^{(i)}_f)\}_{i=0}^h,
\]
where $\mathcal{CC}^0=\mathcal{CC}$ and $\ell_0=(\ell^{(0)}_v,\ell^{(0)}_e,\ell^{(0)}_f)$.
\end{Definition}

CCWL test employs by labeling the vertices, edges and faces 
with $\ell^{(0)}_v,\ell^{(0)}_e,\ell^{(0)}_f=0$. For iteration $k=0,\dots,h-1$, $\ell^{(k+1)}$ are relabeled with the $k$-th labels $\ell^{(k)}$ on edge, face and nodes, respectively. The new labels are updated as
\begin{equation}
\small
\begin{aligned}
\ell_v^{(k+1)} & = \{\!\{
\ell_v^{(k)},
\{\!\{\ell_e^{(k)}\}\!\}_{e \in E_v}
\}\!\}. \\
\ell_e^{(k+1)} & = \{\!\{
\ell_e^{(k)},
\{\!\{\ell_v^{(k)}\}\!\}_{v \in V_e},
\{\!\{\ell_f^{(k)}\}\!\}_{f \in F_e}
\}\!\},
 \\
\ell_f^{(k+1)} & = \{\!\{
\ell_f^{(k)},
\{\!\{\ell_e^{(k)}\}\!\}_{e \in E_f} \}\!\},
\end{aligned}
\end{equation}
where $E_v, V_e,F_e,E_f$ are the edges connected to nodes, nodes connected to edges, faces connected to edges and edges connected faces, respectively. $\{\{\}\}$ denotes a multiset.

1-$\mathcal{CC}$WL distinguish $CC_1$ and $CC_2$ as non-isomorphic if there exists a cell at $k$-iteration such that the condition
\begin{equation}
\label{condition01}
 \{\!\{ l_{\mathcal{CC}_1,\sigma_1}^{(k)} | \sigma_1 \in \mathcal{CC}_1 \}\!\} \ne \{\!\{ l_{\mathcal{CC}_2,\sigma_2}^{(k)} | \sigma_2 \in \mathcal{CC}_2 \}\!\}, 
\end{equation}
then we can conclude that
\begin{Proposition}[(1-CCWL)]
\label{proposition01}
If 1-CCWL test decides $\mathcal{CC}_1$ and $\mathcal{CC}_2$ are non-isomorphic, then $\mathcal{CC}_1 \ne \mathcal{CC}_2$.
\end{Proposition}

\begin{proof}[\textbf{Proof for Proposition \ref{proposition01}}]
We prove the contrapositive. Assume that $\mathcal{CC}_1$ and $\mathcal{CC}_2$ are isomorphic
combinatorial complexes.
Let $\mathcal{CC}_1 = (\mathcal{S}_1,\mathcal{C}_1,\mathrm{rk})$ and
$\mathcal{CC}_2 = (\mathcal{S}_2,\mathcal{C}_2,\mathrm{rk})$.
By isomorphism definition, there exists an injection function 
$\varphi:\mathcal{CC}_1 \to \mathcal{CC}_2$.
For all
$\sigma \in \mathcal{CC}_1$, the ranks are preserved: $\mathrm{rk}(\sigma) = \mathrm{rk}(\varphi(\sigma))$ (i.e., nodes map to nodes, edges to edges, faces to faces). 
Moreover, the incidence matrix is preserved: for each cell $\sigma \in \mathcal{CC}_1$, the sets of
incident nodes, edges and faces correspond under $\varphi$. It holds 
$\varphi(E_v) = E_{\varphi(v)},$ $
\varphi(V_e) = V_{\varphi(e)},$ $
\varphi(F_e) = F_{\varphi(e)}$,
$\varphi(E_f) = E_{\varphi(f)}$. 
We show by induction on $k$ that the labels produced by the 1-CCWL refinement are preserved
under $\varphi$ at every iteration, i.e., for all $\sigma \in \mathcal{CC}_1,\ k \ge 0.$
, there holds $
\ell^{(k)}_{\mathcal{CC}_2,\varphi(\sigma)} = \ell^{(k)}_{\mathcal{CC}_1,\sigma}$.

At $k=0$, By definition, all nodes, edges and faces are initialized with the same labels
$\ell_v^{(0)},\ell_e^{(0)},\ell_f^{(0)} = 0$ in both complexes, so that $\ell^{(0)}_{\mathcal{CC}_2,\varphi(\sigma)} = \ell^{(0)}_{\mathcal{CC}_1,\sigma}$. 
For some $k \ge 0$, we suppose that labels at iteration $k$ are preserved under $\varphi$. Then we conclude that it also holds for iteration $k+1$ for nodes, edges and faces. 
We employ the 1-CCWL update rule on two combinatorial complexes.
For the node $v \in \mathcal{CC}_1$. we have
$
\ell^{(k+1)}_{v}
=
\{\!\{
\ell^{(k)}_{v},
\{\!\{\ell^{(k)}_{e}\}\!\}_{e \in E_v}
\}\!\}.
$ 
Since $\varphi$ is an isomorphism preserving incidence and, by the induction hypothesis,
$\ell^{(k)}_{\mathcal{CC}_2,\varphi(x)} = \ell^{(k)}_{\mathcal{CC}_1,x}$ for all cells $x$,
the multiset
$\{\!\{\ell^{(k)}_{e}\}\!\}_{e \in E_v}$ in $\mathcal{CC}_1$ coincides with
$\{\!\{\ell^{(k)}_{e'}\}\!\}_{e' \in E_{\varphi(v)}}$ in $\mathcal{CC}_2$.
Hence
$
\ell^{(k+1)}_{\mathcal{CC}_2,\varphi(v)}
=
\ell^{(k+1)}_{\mathcal{CC}_1,v}$ .
The same argument applies to edges and faces. 
For an edge $e$, 
$ 
\ell_e^{(k+1)}
=
\{\!\{
\ell_e^{(k)},
\{\!\{\ell_v^{(k)}\}\!\}_{v \in V_e},
\{\!\{\ell_f^{(k)}\}\!\}_{f \in F_e}
\}\!\}$, 
and the isomorphism $\varphi$ preserves both $V_e$ and $F_e$ as well as all labels at 
iteration $k$, so
$
\ell^{(k+1)}_{\mathcal{CC}_2,\varphi(e)}
=
\ell^{(k+1)}_{\mathcal{CC}_1,e}$. For a face $f$, we also have
$
\ell_f^{(k+1)}
=
\{\!\{
\ell_f^{(k)},
\{\!\{\ell_e^{(k)}\}\!\}_{e \in E_f}
\}\!\},
$
and again the incidence-preserving property of $\varphi$ and the induction hypothesis imply
$\ell^{(k+1)}_{\mathcal{CC}_2,\varphi(f)}
=\ell^{(k+1)}_{\mathcal{CC}_1,f}$. 
Thus, by induction, for every iteration $k$ and every cell $\sigma \in \mathcal{C}_1$, the labels
of $\sigma$ and $\varphi(\sigma)$ coincide. In particular, the multisets of labels in
$\mathcal{CC}_1$ and $\mathcal{CC}_2$ are identical at each iteration $k$:
\[
\{\!\{ \ell^{(k)}_{\mathcal{CC}_1,\sigma_1} : \sigma_1 \in \mathcal{CC}_1 \}\!\}
=
\{\!\{ \ell^{(k)}_{\mathcal{CC}_2,\sigma_2} : \sigma_2 \in \mathcal{CC}_2 \}\!\}.
\]
Therefore, the 1-CCWL test cannot distinguish isomorphic combinatorial complexes.
Taking the contrapositive, if the 1-CCWL test distinguishes $\mathcal{CC}_1$ and $\mathcal{CC}_2$, there exists a cell at $k$-iteration such that $ \{\!\{ l_{\mathcal{CC}_1,\sigma_1}^{(k)} | \sigma_1 \in \mathcal{CC}_1 \}\!\} \ne \{\!\{ l_{\mathcal{CC}_2,\sigma_2}^{(k)} | \sigma_2 \in \mathcal{CC}_2 \}\!\} $
, then
$\mathcal{CC}_1$ and $\mathcal{CC}_2$ are not isomorphic.
\end{proof}

We can conclude that 
1-$\mathcal{CC}$WL distinguish $CC_1$ and $CC_2$ as non-isomorphic if there exists a cell at $k$-iteration that holds the condition \eqref{condition01}. 
Suppose the same features to all cells in combinatorial complex which allows CCMamba to rely only on the structure to learn. Let $\mathcal{A}:\mathcal{CC}\to R^d$ 
be a CCMamba Layer in Equation \ref{ccmpnn}, we can conclude the Lemma as 
\begin{Lemma} 
\label{lemma1}
Suppose an injective function $\mathcal{A}: \mathcal{CC}_1\to \mathcal{CC}_2$, and let the labels $\ell_{\mathcal{CC}}^{(k)}=\{\ell^{(k)}_v, \ell^{(k)}_e, \ell^{(k)}_f \}$, and features $h_{\mathcal{CC}}^{(k)}=\{h^{(k)}_v, h^{(k)}_e, h^{(k)}_f\}$ of node, edge and face in combinatorial complex $\mathcal{CC}$ at iteration $k$, respectively. If the conditions hold, 
\begin{equation}
\small
\begin{cases}
l_{\mathcal{CC}_1,\sigma_1}^{(j)} = l_{\mathcal{CC}_2,\sigma_2}^{(j)},& \forall j \le k \\
h_{\mathcal{CC}_1,\sigma_1}^{(j)} = h_{\mathcal{CC}_2,\sigma_2}^{(j)},& \forall j \le k-1
\end{cases}
\end{equation}
for iterations $0,1,\dots,k$. Then, for the condition in Equation \ref{condition01} holds, it follows that $ h_{\mathcal{CC}_1,\sigma_1}^{(k)}= h_{\mathcal{CC}_2,\sigma_2}^{(k)}$ at iterations $ k $ .
\end{Lemma}

\begin{proof}[\textbf{Proof of Lemma \ref{lemma1}}]
We prove this condition by induction on the iteration $k$.
At iteration $k=0$, all cells are initialized with the same feature vector. By assumption,
the initial labels coincide, then the statement holds for $k=0$.

Suppose that the claim holds for iterations $\forall j \le k-1$, i.e., for any pair of cells
$\sigma_1 \in \mathcal{CC}_1$, $\sigma_2 \in \mathcal{CC}_2$ with
$\ell_{\mathcal{CC}_1,\sigma_1}^{(j)} = \ell_{\mathcal{CC}_2,\sigma_2}^{(j)}$. In the case of the function $\mathcal{A}$ is an injective function, 
we also have
$h_{\mathcal{CC}_1,\sigma_1}^{(j)} = h_{\mathcal{CC}_2,\sigma_2}^{(j)}$, where the cells $\sigma_1,\sigma_2$ can be nodes (0-cells), edges (1-cells),and faces (2-cells).

Let $E_v$ be the set of incident edges of a node $v$,
$V_e, F_e$ the incident nodes and faces of an edge $e$, and $E_f$ the incident edges of a face $f$.
For an iteration $k$ and a pair of cells $\sigma_1 \in \mathcal{CC}_1$, $\sigma_2 \in \mathcal{CC}_2$
such that
$\ell_{\mathcal{CC}_1,\sigma_1}^{(k)} = \ell_{\mathcal{CC}_2,\sigma_2}^{(k)}$.
By the definition of the 1-CCWL update rule, this implies that:
\begin{equation}
\begin{aligned}
 \underset{v \in E_v(\sigma_1)}{\{\!\{\ell_v^{(k-1)}\}\!\}},\ 
 \underset{e \in V_e(\sigma_1)\cup F_e(\sigma_1)}{\{\!\{\ell_e^{(k-1)}\}\!\}},\ 
 \underset{f \in E_f(\sigma_1)}{\{\!\{\ell_f^{(k-1)}\}\!\}} \\
\;=\;\ 
 \underset{v \in E_v(\sigma_2)}{\{\!\{\ell_v^{(k-1)}\}\!\}},\ 
 \underset{e \in V_e(\sigma_2)\cup F_e(\sigma_2)}{\{\!\{\ell_e^{(k-1)}\}\!\}},\ 
 \underset{f \in E_f(\sigma_2)}{\{\!\{\ell_f^{(k-1)}\}\!\}}. 
 \end{aligned}
\end{equation}

It indicates that at iteration $k-1$ the multisets of neighbor labels around $\sigma_1$ and $\sigma_2$ coincide.
By the inductive hypothesis, equality of labels up to iteration $k-1$ implies equality of features
up to iteration $k-1$.
Consequently, the above multiset equalities also hold at the feature level. 
\begin{equation}
\begin{aligned}
 \underset{v \in E_v(\sigma_1)}{\{\!\{h_v^{(k-1)}\}\!\}},\ 
 \underset{e \in V_e(\sigma_1)\cup F_e(\sigma_1)}{\{\!\{h_e^{(k-1)}\}\!\}},\ 
 \underset{f \in E_f(\sigma_1)}{\{\!\{h_f^{(k-1)}\}\!\}} \\
\;=\;
 \underset{v \in E_v(\sigma_2)}{\{\!\{h_v^{(k-1)}\}\!\}},\ 
 \underset{e \in V_e(\sigma_2)\cup F_e(\sigma_2)}{\{\!\{h_e^{(k-1)}\}\!\}},\ 
 \underset{f \in E_f(\sigma_2)}{\{\!\{h_f^{(k-1)}\}\!\}}.
 \end{aligned}
\end{equation}

By construction, the CCMamba layer $\mathcal{A}$ updates $h^{(k)}_\sigma$ through
injective projection and permutation-invariant aggregation over these neighbor-feature multisets
(at cell level) together with the previous feature of $\sigma$ itself.
Since all these inputs coincide for $\sigma_1$ and $\sigma_2$ at iteration $k-1$,
the injectivity of $\mathcal{A}$ implies
$
h^{(k)}_{\mathcal{CC}_1,\sigma_1}
=
h^{(k)}_{\mathcal{CC}_2,\sigma_2}.
$
\end{proof}

We further conclude that the upper bounded of CCMamba layer $\mathcal{A}$'s expressive power is 1-CCWL test. 

\begin{Proposition}
\label{proposition02}
Given two non-isomorphic combinatorial complexes $\mathcal{CC}_1$ and $\mathcal{CC}_2$, if $\mathcal{A}$ can distinguish them by $\mathcal{A}(\mathcal{CC}_1) \ne \mathcal{A}(\mathcal{CC}_2)$, then 1-CCWL test also decides $\mathcal{CC}_1 \ne \mathcal{CC}_2 $.
\end{Proposition}

\begin{proof}[\textbf{Proof of Proposition \ref{proposition02}}]
For the condition that combinatorial complexes $\mathcal{CC}_1$ and $\mathcal{CC}_2$ are non-isomorphic, and $\mathcal{A}$ can distinguish them by $\mathcal{A}(\mathcal{CC}_1) \ne \mathcal{A}(\mathcal{CC}_2)$. 
Let the labels $\ell_{\mathcal{CC}}^{(k)}=\{\ell^{(k)}_v, \ell^{(k)}_e, \ell^{(k)}_f \}$, and features $h_{\mathcal{CC}}^{(k)}=\{h^{(k)}_v, h^{(k)}_e, h^{(k)}_f\}$ of node, edge and face in combinatorial complex $\mathcal{CC}$ at iteration $k$, respectively.
We obtain that there exists $k$ satisfies that: 
\begin{equation}
\label{eq05}
\begin{cases}
l_{\mathcal{CC}_1,\sigma_1}^{(j)} = l_{\mathcal{CC}_2,\sigma_2}^{(j)},
h_{\mathcal{CC}_1,\sigma_1}^{(j)} = h_{\mathcal{CC}_2,\sigma_2}^{(j)}, \forall j \le k-1 \\
h_{\mathcal{CC}_1,\sigma_1}^{(k)} \ne h_{\mathcal{CC}_2,\sigma_2}^{(k)}
\end{cases}
\end{equation}

If we suppose that 1-CCWL test cannot distinguish $\mathcal{CC}_1, \mathcal{CC}_2 $, It indicates that at iteration $k$ holds 
\begin{equation}
\label{eq06}
\begin{aligned}
 \underset{v \in E_v(\sigma_1)}{\{\!\{\ell_v^{(k)}\}\!\}},\ 
 \underset{e \in V_e(\sigma_1)\cup F_e(\sigma_1)}{\{\!\{\ell_e^{(k)}\}\!\}},\ 
 \underset{f \in E_f(\sigma_1)}{\{\!\{\ell_f^{(k)}\}\!\}} \\
\;=\;\ 
 \underset{v \in E_v(\sigma_2)}{\{\!\{\ell_v^{(k)}\}\!\}},\ 
 \underset{e \in V_e(\sigma_2)\cup F_e(\sigma_2)}{\{\!\{\ell_e^{(k)}\}\!\}},\ 
 \underset{f \in E_f(\sigma_2)}{\{\!\{\ell_f^{(k)}\}\!\}}, 
 \end{aligned}
\end{equation}
which concludes that $l_{\mathcal{CC}_1,\sigma_1}^{(k)} = l_{\mathcal{CC}_2,\sigma_2}^{(k)}$. According to the Lemma \ref{lemma1}, $h_{\mathcal{CC}_1,\sigma_1}^{(k)}= h_{\mathcal{CC}_2,\sigma_2}^{(k)}$. This contradicts the assumption of formula \ref{eq06}. Thus, the proof of the Proposition finished.

\end{proof}

\begin{Theorem}
\label{theorem}
 Given two non-isomorphic combinatorial complexes $\mathcal{CC}_1$ and $\mathcal{CC}_2$, a CCMamba Layer $\mathcal{A}$ can distinguish them, if conditions hold. Under the conditions of Lemma~\ref{lemma1} and with an injective permutation-invariant readout

1. Local Level. $\phi_1,\phi_2$ functions are injective.

2. Global Level. In addition to the local-level conditions, $\mathcal{A}$’s graph-level READOUT function is injective.
\end{Theorem}

\begin{proof}[\textbf{Proof of Theorem~\ref{theorem}}]
By Proposition~\ref{proposition02}, the distinguishing power of any CCMamba layer $\mathcal{A}$ is
{upper-bounded} by 1-CCWL test. If a CCMamba layer $\mathcal{A}$ distinguishes
$\mathcal{CC}_1$ and $\mathcal{CC}_2$, then 1-CCWL distinguishes them as well.
It remains to prove the converse: if 1-CCWL distinguishes two complexes, then there exists a
CCMamba layer (satisfying the stated injectivity assumptions) that distinguishes them.

We first demonstrate that the CCMamba layer can fit the mapping of the 1-CCWL test. 
Assume 1-CCWL distinguishes $\mathcal{CC}_1$ and $\mathcal{CC}_2$, there exists an iteration $k$ such that the multisets of 1-CCWL labels are not equal:
\begin{equation}\label{eq:ccwl_diff}
\{\!\{ \ell^{(k)}_{\mathcal{CC}_1,\sigma} : \sigma_1 \in \mathcal{CC}_1 \}\!\}
\neq
\{\!\{ \ell^{(k)}_{\mathcal{CC}_2,\sigma} : \sigma_2 \in \mathcal{CC}_2 \}\!\}.
\end{equation}

By induction on $k\in\{0,\dots,K\}$, that one can parameterize the CCMamba layers
$\mathcal{A}^{(0)},\dots,\mathcal{A}^{(k-1)}$ such that at each depth $k$ there exists an
injective map $\Phi^{(k)}$ (i.e., summation function) with for $\forall \sigma \in \mathcal{CC}$, we have
\begin{equation}\label{eq:phi_k}
h^{(k)}_{\mathcal{CC},\sigma}=\Phi^{(k)}\!\big(\ell^{(k)}_{\mathcal{CC},\sigma}\big),
\end{equation}
which means the feature $h^{(k)}_{\sigma}$ is an injective encoding of the 1-CCWL label
$\ell^{(k)}_{\sigma}$.
At initialization $k=0$, all cells of a fixed rank initials the same label $\ell^{(0)}_{\sigma}$, and the layers initializes all corresponding features $h^{(0)}_{\sigma}$identically. We can conclude the condition \eqref{eq:phi_k} holds for
$k=0$ by defining $\Phi^{(0)}$ to map the (unique) initial label to the initial features.

Assume \eqref{eq:phi_k} holds for some $k\ge 0$ with an injective $\Phi^{(k)}$, 
We prove that there exists a parameterization of $\mathcal{A}^{(k)}$ and an injective
$\Phi^{(k+1)}$ such that the condition \eqref{eq:phi_k} also holds at iteration $k+1$. 
From the perspective of rank definition, we consider cells of different ranks, including nodes, edges, and faces. Within the 1-CCWL, the refined label $\ell^{(k+1)}_{\sigma}$ is obtained
by applying an injective map function to the previous label of nodes $v$, edges $e$, faces $f$, and 
the multisets of labels of incident neighbors, grouped by rank, as follows
\begin{equation}
 \small
\begin{aligned}
\ell_v^{(k+1)} & = \mathrm{HASH}\Big(\ell_v^{(k)},
\{\!\{\ell_e^{(k)}\}\!\}_{e \in E_v}\Big), \\
\ell_e^{(k+1)} & = \mathrm{HASH}\Big(\ell_e^{(k)},
\{\!\{\ell_v^{(k)}\}\!\}_{v \in V_e},
\{\!\{\ell_f^{(k)}\}\!\}_{f \in F_e}\Big), \\
\ell_f^{(k+1)} & = \mathrm{HASH}\Big(\ell_f^{(k)},
\{\!\{\ell_e^{(k)}\}\!\}_{e \in E_f}\Big).
\end{aligned}
\end{equation}

We define the rank neighbor multiset operator
$\mathcal{N}_{\tau\rightarrow \sigma,r}^{(k)}(\sigma)$ to be the tuple consisting of the multisets
$\{\!\{\ell^{(k)}_{\tau}\}\!\}_{\tau\in\mathcal{N}_r(\sigma)}$ for each relevant rank $r$.
Then there exists an injective function $\eta$ ( HASH map), it satisfies
\begin{equation}
\label{eq:ccwl_update_abstract}
\ell^{(k+1)}_{\sigma}=\eta\big(\ell^{(k)}_{\sigma},\,\mathcal{N}_{\tau\rightarrow \sigma,r}^{(k)}(\sigma)\big).
\end{equation}

By the inductive hypothesis, $\Phi^{(k)}$ is injective, hence it induces a bijection between
labels appearing at depth $k$ and their encoded features. For every $\sigma$, the pair
$\big(\ell^{(k)}_{\sigma},\mathcal{N}_{\tau\rightarrow \sigma ,r }^{(k)}(\sigma)\big)$ correspond to the pair $\big(h^{(k)}_{\sigma},\hat{\mathcal{N}}_{\tau\rightarrow \sigma}^{(k)}(\sigma)\big)$, where
$\hat{\mathcal{N}}_{\tau\rightarrow \sigma}^{(k)}(\sigma)$ is obtained by replacing each neighbor label multiset by
the multiset of the corresponding neighbor features.

From local-level assumptions, the CCMamba update $\mathcal{A}^{(k)}$ 
(i) $\phi_1$ maps injectively the per-rank aggregations, and (ii) $\phi_2$ combines the fused neighbor summation function with the center feature. Therefore, there exists a update map 
$(h^{(k)}_{\sigma},\hat{\mathcal{N}}_{\tau\rightarrow \sigma}^{(k)}(\sigma))
\longmapsto h^{(k+1)}_{\sigma}$ 
is injective as a function of the information in 
$\big(\ell^{(k)}_{\sigma},\hat{\mathcal{N}}_{\tau\rightarrow \sigma}^{(k)}(\sigma)\big)$.
Combining this with \eqref{eq:ccwl_update_abstract}, we can define an injective map
$\Phi^{(k+1)}$ on labels by setting
$
\Phi^{(k+1)}\!\big(\ell^{(k+1)}_{\sigma}\big)\;:=\;h^{(k+1)}_{\sigma}$, 
which is injective because $h^{(k+1)}_{\sigma}$ is an injective function of
$\ell^{(k+1)}_{\sigma}$.
Thus the condition \eqref{eq:phi_k} holds at depth $k+1$. 
By induction, \eqref{eq:phi_k} holds for all $k=0,\dots,K$.

Furthermore, we prove the distinguishability of the CCMamba Layer.
Considering the specific CCMamba layer constructed with $h^{(k)}_{\mathcal{CC},\sigma}=\Phi^{(k)}\!\big(\ell^{(k)}_{\mathcal{CC},\sigma}\big)$ up to
depth $K$). Suppose that this layer $\Phi^{(k)}$ cannot distinguish
$\mathcal{CC}_1$ and $\mathcal{CC}_2$. Then for $\sigma_1\in\mathcal{CC}_1$,$\sigma_2 \in \mathcal{CC}_2$, the final READOUT outputs:
\begin{equation}
\small
\begin{aligned}
\label{eq:readout_equal}
\mathrm{READOUT}\big(\{\!\{h^{(T)}_{\mathcal{CC}_1,\sigma_1}\}\!\}\big) 
=
\mathrm{READOUT}\big(\{\!\{h^{(T)}_{\mathcal{CC}_2,\sigma_2}\}\!\}\big).
\end{aligned}
\end{equation}
By the global-level assumption, READOUT is injective over multisets of cell features, the condition
\eqref{eq:readout_equal} implies the input multisets of features are equal as 
\begin{equation}\label{eq:feat_multiset_equal}
\{\!\{h^{(T)}_{\mathcal{CC}_1,\sigma}:\sigma_1\in\mathcal{CC}_1\}\!\}
=
\{\!\{h^{(T)}_{\mathcal{CC}_2,\sigma}:\sigma_2\in\mathcal{CC}_2\}\!\}.
\end{equation}
Since $\Phi^{(T)}$ is injective and $h^{(T)}_{\sigma}=\Phi^{(T)}(\ell^{(T)}_{\sigma})$, equality of
feature multisets in \eqref{eq:feat_multiset_equal} implies equality of label multisets:
\[
\{\!\{\ell^{(T)}_{\mathcal{CC}_1,\sigma}:\sigma\in\mathcal{CC}_1\}\!\}
=
\{\!\{\ell^{(T)}_{\mathcal{CC}_2,\sigma}:\sigma\in\mathcal{CC}_2\}\!\},
\]
contradicting \eqref{eq:ccwl_diff}. Therefore, the constructed CCMamba layer must distinguish
$\mathcal{CC}_1$ and $\mathcal{CC}_2$. we have proved CCMamba cannot distinguish more pairs than 1-CCWL
(Proposition~\ref{proposition02}), and whenever 1-CCWL distinguishes two complexes, there
exists a CCMamba layer satisfied the injectivity assumptions  that distinguishes them.
Hence, under this conditions, CCMamba and 1-CCWL have exactly the same distinguishing
power.
\end{proof}


\section{Experiments}
In this section, we evaluate the performance of the proposed framework CCMamba in extensive experiments.

\subsection{Experimental Setup}

\paragraph{Datasets}
We evaluate CCMamba on the real-world benchmarks. 
For {graph classification}, we use standard TUDatasets~\cite{morris2020tudataset}, including:
molecular graphs (MUTAG, PROTEINS) for chemical property and protein function prediction; social networks (IMDB-BINARY, IMDB-MULTI); and heterophilic graphs from the HeterophilicGraphDataset~\cite{platonovcritical}, namely AMAZON-RATINGS, ROMAN-EMPIRE, and MINESWEEPER.
At the node classification level, we adopt the citation networks Cora, CiteSeer, and PubMed~\cite{yang2016revisiting}. Following the structural lifting methods provided by \textsc{TopoBench}~\cite{telyatnikov2024topobenchmarkx}, we lift all base graphs into higher-order structures for fair comparison with the baseline methods, such as hypergraphs, simplicial complexes, and cellular complexes. The dataset statistics are summarized in Table~\ref{datasets}.

\begin{table}[htbp]
\setlength{\tabcolsep}{2pt}
\caption{The statistics of datasets.}
\centering
\begin{tabular}{lccccc} 
 \toprule 
 Datasets & Nodes & Edges & Features & Classes & Tasks \\
 \midrule 
 MUTAG & 3371 & 3371 & 7 & 2 & graph classification \\
 PROTEINS & 43471 & 81044 & 3 & 2 & graph classification \\
 IMDB-BINARY & 19770 & 193100 & 0 & 2 & graph classification \\
 IMDB-MULTI & 19500 & 98910 & 0 & 3 & graph classification \\
 AMAZON-RATINGS & 24492 & 93050 & 300 & 5 & graph classification \\
 ROMAN-EMPIRE & 22662 & 32927 & 300 & 18 & graph classification \\
 MINESWEEPER & 10000 & 39402 & 7 & 2 & graph classification \\
 Cora & 2708 & 10556 & 1433 & 7 & node classification \\
 CiteSeer & 3327 & 9104 & 3703 & 6 & node classification \\
 PubMed & 19717 & 88648 & 500 & 3 & node classification \\
 \bottomrule 
\end{tabular}
\label{datasets}
\end{table}

\begin{itemize}
 \item \textbf{MUTAG} dataset is a collection of nitroaromatic compounds, where nodes represent atoms labelled by atomic type, and edges denote the bridges between atoms.
 
 \item \textbf{PROTEINS}: This dataset comprises proteins classified as either enzymatic or non-enzymatic. Nodes represent amino acids, and if the distance between two nodes is less than 6 $\AA$ ($1\AA=10^{-10}$m), they are connected.
 
 \item \textbf{IMDB-BINARY} \& \textbf{IMDB-MULTI}: These two datasets are film collaboration datasets, comprising self-directed networks of actors/actresses who have roles in films listed on IMDB. Nodes represent actors/actresses, while edges denote their appearances in the same film. In IMDB-BINARY, edges are drawn from action and romantic films, and in IMDB-MULTI, edges are drawn from comedies, romantic films, and science fiction films.
 
 \item \textbf{AMAZON-RATINGS}: This dataset consists of Amazon product co-purchase network metadata, where nodes represent products and edges denote items frequently purchased together.

 \item \textbf{ROMAN-EMPIRE}: This dataset is based on the English Wikipedia article Roman Empire, where a node represents a word in the text. An edge is established if two words appear consecutively in the text or are connected in the dependency tree of a sentence.

 \item \textbf{MINESWEEPER}: The node of this dataset corresponds to a grid square in the Minesweeper game. Each square connects to its eight neighbours, forming a set of edges.

 \item \textbf{Cora},\textbf{Citeseer}, and \textbf{PubMed}: These three datasets serve as benchmark datasets for node classification. Each node represents a document, while edges denote citation relationships between documents.


 
\end{itemize}

\begin{table*}[htbp]
\centering
\caption{Detailed hyperparameter settings of CCMamba on different datasets.}
\label{tabparams}
\setlength{\tabcolsep}{3pt} 
\begin{tabular}{lcccccccccccc}
\toprule
\multirow{2}{*}{Dataset} & \multicolumn{11}{c}{CCMamba Hyperparameters} \\
\cmidrule{2-12}
 & \makecell{Batch\\ Size} & \makecell{Learning \\ Rate} & \makecell{Maximum Cells\\ Length} & \makecell{Random \\ Seeds} & \makecell{Hidden \\ Dimension} & Layers & \makecell{SSM State  \\ Dimension} & Expand & Dropout & Epochs & Input Ranks \\
\midrule
IMDB-BINARY & 256 & 0.005 & 10 & 0,3,5,7,9 & 128 & 3 & 16 & 2 & 0.5 & 500 & [0, 1, 2] \\
IMDB-MULTI & 256 & 0.01 & 10 & 0,3,5,7,9 & 64 & 1 & 16 & 2 & 0.5 & 500 & [0, 1] \\
MUTAG & 32 & 0.005 & 15 & 0,3,5,7,9 & 128 & 2 & 16 & 2 & 0.25 & 500 & [0, 1] \\
PROTEINS & 256 & 0.01 & 15 & 0,3,5,7,9 & 64 & 4 & 16 & 2 & 0.25 & 500 & [0, 1] \\
AMAZON\_RATINGS & 1 & 0.005 & 10 & 0,3,5,7,9 & 128 & 3 & 16 & 2 & 0.25 & 1000 & [0, 1] \\
MINESWEEPER & 1 & 0.01 & 10 & 0,3,5,7,9 & 64 & 2 & 16 & 2 & 0.25 & 1000 & [0, 1, 2] \\
ROMAN\_EMPIRE & 1 & 0.01 & 10 & 0,3,5,7,9 & 64 & 2 & 16 & 2 & 0.5 & 1000 & [0, 1, 2] \\
Cora & 1 & 0.005 & 10 & 0,3,5,7,9 & 128 & 2 & 16 & 2 & 0.5 & 1000 & [0, 1, 2] \\
CiteSeer & 1 & 0.005 & 10 & 0,3,5,7,9 & 128 & 2 & 16 & 2 & 0.25 & 1000 & [0, 1, 2] \\
PubMed & 1 & 0.005 & 10 & 0,3,5,7,9 & 64 & 2 & 16 & 2 & 0.5 & 1000 & [0, 1, 2] \\
\bottomrule
\end{tabular}
\end{table*}

\paragraph{Baselines Models} We compare against a broad range of baselines spanning multiple topological domains: 
Graph neural networks, Hypergraph neural networks, Simplicial complex networks and Cellular complexes networks. 
All models are evaluated on the same lifted higher-order structures (hypergraphs, simplicial complexes, or cellular complexes) to ensure a fair comparison. The details of baseline methods as below.


\begin{itemize}
 \item \textbf{Graph Neural Networks:} \textbf{GCN}~\cite{kipf2016semi}: Graph Convolutional Networks that enhance stability via the spectral graph convolutions. \textbf{GAT}~\cite{vaswani2017attention}: Graph Attention Networks that aggregate neighbor features. \textbf{GIN}~\cite{xupowerful}: Graph Isomorphism Networks that employ learnable aggregators to distinguish node and graph tasks. We include them to validate the necessity of higher-order interactions.

 \item \textbf{Hypergraph Neural Networks:} 
 \textbf{HyperGCN}~\cite{hypergcn_neurips19} treats hyperedges as aggregation units and updates node representations through weighted intra-hyperedge feature aggregation and \textbf{HyperGAT}~\cite{ding2020more} extend graph attention to hypergraphs by computing attention scores within hyperedges. 
 \textbf{UniGNN}~\cite{huang2021unignn} (including UniGCN, UniGNNII, UniSAGE, UniGIN) unifies graph and hypergraph message passing. 
 \textbf{AllSet}~\cite{chienyou} integrates deep sets and set Transformers with permutation-invariant aggregators to flexibly model higher-order relations. \textbf{EDGNN}~\cite{jo2021edge} jointly encodes node features, edge labels during neighborhood aggregation for hypergraph representation.

 \item \textbf{Simplicial Neural Networks:} 
 \textbf{SCCN/SCCNN}~\cite{yang2022efficient} learn representations over relaxed simplex complexes without dimensional constraints on simplex-level tasks. \textbf{SCN}~\cite{roddenberry2021principled,lecha2025higher} introduce simplicial convolutions that respect directional invariance. 
 \textbf{SANN}~\cite{gurugubelli2023sann} focuses on computational efficiency via pre-aggregation, while \textbf{SGAT}~\cite{lee2022sgat} utilizes multi-head attention for simplicial interactions.





 \item \textbf{Cellular Neural Networks:} 
 \textbf{CCNN/CXN}~\cite{hajij2020cell} generalize convolutions to cell complexes. 
 \textbf{CWN}~\cite{bodnar2021weisfeiler} enhances expressivity through cellular Weisfeiler-Lehman message passing. 
 \textbf{CCCN}~\cite{papillontopotune} is a foundational topological deep learning model that leverages combinatorial complex representations to capture arbitrary higher-order interactions and hierarchical relationships. 
 
 
\end{itemize}

\paragraph{Implementation Details} 
All experiments are implemented in PyTorch and conducted on two NVIDIA RTX 6000 GPUs (48 GB). All models are trained for up to 1000 epochs and evaluated using 5-fold cross-validation and report the average performance across five independent runs with distinct random seeds $\{0, 3, 5, 7, 9\}$.
Regarding hyperparameter configurations, the hidden dimension for all CCMamba variants is set to 128, while the state dimension $d_{state}$ and expansion factor $E$ of the selective state-space models are set to 16 and 2, respectively. The parameters of CCMamba are set differently for different datasets, as shown in Table \ref{tabparams}.

\paragraph{Performance Metrics} we employ Classification accuracy~\cite{liu2025graph} to evaluate the performance for both node-level and graph-level classification tasks.
Graph classification defines as a set of graphs$\{G_i\}^k_{i=1} \subseteq \mathcal{G}$ with the labels $y_i$. This task aims to learn a function: $f: \mathcal{G} \to \mathcal{Y} $ to predict chemical molecular properties and classify social network groups. Similarly, node classification~\cite{kipf2016semi} denotes as Given a graph $G = (V,E)$, each node $v\in V$ with partial labels $y_v$, and supervised information $\{(v_i,y_i)\}^k_{i=1} \subset V \times Y$, it learns a function $f : V \to Y$ such that $f(v) \approx y_v, \forall v \in V$. This task aims to predict the labels of nodes within social networks and citation networks.

\begin{table*}[ht]
\caption{Test accuracy (\%) of CCMamba and baseline methods across graph classification and node classification on citation networks. Best results (\textbf{bold}) and second-best scores (\underline{underlined}).}
\setlength{\tabcolsep}{2pt}
\centering
\begin{tabular}{c|c|ccccccc|ccc|c}
\toprule 
& Model & \multicolumn{7}{c}{\textbf{Graph Level Tasks}} & \multicolumn{3}{c}{\textbf{Node Level Tasks}} & \multirow{1}{*}{\textbf{Mean}} \\
 \cmidrule{3-9} \cmidrule{9-12} 
& & MUTAG & PROTEINS & IMDB-B & IMDB-M & ROMAN-E & AMAZON-R & Minesweeper & Cora & Citeseer & PubMed & \\
\midrule
\multirow{3}*{\rotatebox{90}{Graph}} & GCN & 70.21$\pm$2.43 & 72.76$\pm$1.78 & 70.80$\pm$2.04 & 49.33$\pm$4.52 & 78.10$\pm$1.15 & 49.84$\pm$3.71 & 83.48$\pm$0.69 & 87.00$\pm$2.11 & 75.51$\pm$3.47 & 89.48$\pm$1.29 & 72.65 \\
& GAT & 72.34$\pm$1.69 & 74.91$\pm$2.34 & 68.40$\pm$3.73 & 46.93$\pm$4.25 & 83.25$\pm$1.82 & 49.29$\pm$3.94 & 83.84$\pm$1.01 & 86.71$\pm$2.25 & 73.59$\pm$3.77 & 89.21$\pm$1.25 & 72.75\\
& GIN & 78.72$\pm$2.61 & 76.70$\pm$1.87 & 71.20$\pm$1.86 & 46.40$\pm$4.07 & 79.92$\pm$2.20 & 49.22$\pm$3.58 & 83.60$\pm$0.37 & 87.59$\pm$1.92 & 73.11$\pm$4.08 & 89.51$\pm$0.94 &73.18 \\
\midrule
\multirow{8}*{\rotatebox{90}{Hypergraph}}& HperGCN & 75.53$\pm$2.05 & 73.82$\pm$2.17 & 67.60$\pm$2.38 & 45.87$\pm$4.18 & 76.40$\pm$1.92 & 48.91$\pm$3.66 & 80.24$\pm$1.27 & 86.35$\pm$4.26 & 72.41$\pm$2.04 & 88.35$\pm$0.97 & 71.55\\
& UniGCN & 72.34$\pm$2.28 & 75.27$\pm$2.06 & 69.20$\pm$3.67 & 47.20$\pm$3.93 & 75.80$\pm$2.17 & 47.46$\pm$4.16 & 80.68$\pm$1.63 & 87.44$\pm$2.38 & 71.31$\pm$4.33 & 85.15$\pm$2.60 &71.19 \\
& HyperGAT & 68.51$\pm$12.04 & 70.25$\pm$2.08 & 70.08$\pm$2.38 & 34.67$\pm$2.08& 71.56$\pm$0.39 & 48.11$\pm$0.16 & 79.82$\pm$0.49 & 85.05$\pm$1.03 & 72.45$\pm$1.83 & 82.87$\pm$0.39 & 68.34\\
& UniGNNII & \underline{82.98$\pm$1.46} & 73.48$\pm$2.68 & 71.20$\pm$2.03 & 47.20$\pm$3.72 & 73.00$\pm$2.98 & 48.75$\pm$3.50 & 80.72$\pm$1.55 & 87.59$\pm$1.73 & 74.79$\pm$3.08 & 88.59$\pm$0.77 & 72.93 \\
& UniGIN & 79.23$\pm$4.81 & 71.26$\pm$2.09 & 68.48$\pm$2.17 & 32.69$\pm$1.06 & 73.29$\pm$0.28 & 48.75$\pm$0.43 & 80.36$\pm$0.79 &84.64$\pm$0.78 & 72.53$\pm$1.09 & 83.08$\pm$0.41 & 69.43 \\
& UniSAGE &78.72$\pm$1.43 & 74.55$\pm$3.76 & 71.60$\pm$4.72 & 34.67$\pm$2.74 & 74.53$\pm$3.81 & 49.70$\pm$0.48 & 80.85$\pm$0.62 & 85.55$\pm$0.76 & 73.78$\pm$1.39 & 83.08$\pm$0.41 & 70.70 \\
& AllSet & 74.47$\pm$2.55 & 76.70$\pm$1.97 & 64.40$\pm$2.92 & 47.47$\pm$3.81 & 77.52$\pm$2.38 & 50.02$\pm$3.27 & 81.88$\pm$1.49 & 83.60$\pm$2.83 & 74.31$\pm$3.29 & 89.05$\pm$1.75 &71.94 \\
& EdGNN & 80.85$\pm$1.63 & 74.91$\pm$2.26 & 70.40$\pm$2.48 & 49.53$\pm$4.15 & 79.99$\pm$1.78 & 48.38$\pm$3.85 & 83.12$\pm$0.99 & 84.44$\pm$2.04 & 75.03$\pm$3.19 & 89.39$\pm$1.15 & 73.60\\
\midrule
\multirow{4}*{\rotatebox{90}{Simplex}}& SCCN & 76.60$\pm$2.06& 75.63$\pm$1.88 & 69.60$\pm$2.84 & 48.20$\pm$3.43 & 83.06$\pm$1.07 & 50.15$\pm$3.49 & 83.92$\pm$0.75 & 81.24$\pm$3.57 & 69.03$\pm$4.55 & 87.85$\pm$1.92 &
72.53 \\
& SAN & 71.38$\pm$5.73 & 71.54$\pm$2.20 & 64.25$\pm$4.27 & 45.35$\pm$2.83 & 84.70$\pm$2.79& 48.99$\pm$3.41 & 82.84$\pm$2.41 & 66.23$\pm$0.71 & 62.50$\pm$0.94 & 88.74$\pm$2.69 & 68.65\\
& SCCNN & 70.21$\pm$2.82 & 75.63$\pm$1.97 & 69.20$\pm$3.04 & 47.73$\pm$3.58 & 84.69$\pm$0.98 & 50.33$\pm$2.91 & 83.36$\pm$0.86 & 82.13$\pm$3.23 & 70.59$\pm$4.28 & 87.83$\pm$2.08 &72.17 \\
& SCN & 72.34$\pm$2.55 & 75.60$\pm$2.07 & 68.80$\pm$2.74 & 47.00$\pm$3.82 & 84.16$\pm$1.13 & 49.88$\pm$3.23 & \underline{85.44$\pm$0.68} & 81.39$\pm$3.09 & 71.91$\pm$4.17 & 88.70$\pm$1.62 
& 72.60\\
& SANN & 74.29$\pm$1.63 & 77.62$\pm$2.28 & 72.7$\pm$2.16 & 48.97$\pm$1.02 & 84.01 $\pm$2.06 & \underline{51.34$\pm$1.72} & 84.73$\pm$0.93 & 82.46$\pm$2.79 & 73.19$\pm$2.11 & 87.26$\pm$0.98 & 73.66
\\
\midrule
\multirow{4}*{\rotatebox{90}{Cellular}}& CCNN & 74.47$\pm$2.17 & 72.04$\pm$2.76 & 72.00$\pm$1.92 & 49.33$\pm$3.27 & 81.54$\pm$1.33 & 51.07$\pm$2.97 & 84.08$\pm$0.83 & 87.44$\pm$2.28 & 74.07$\pm$3.32 & 88.09$\pm$2.15 & 73.41\\
& CCXN & 76.60$\pm$1.92 & 73.48$\pm$2.57 & 70.80$\pm$2.37 & 46.93$\pm$3.83 & 81.15$\pm$1.46 & 48.24$\pm$3.69 & 83.28$\pm$1.08 & 87.30$\pm$2.44 & 73.95$\pm$3.63 & 88.34$\pm$1.87 & 73.01 \\
& CWN & 82.47$\pm$1.34 & 74.19$\pm$2.27 & 72.40$\pm$2.17 & 49.07$\pm$3.42 & 82.55$\pm$1.27 & 50.99$\pm$3.14 & 84.00$\pm$0.73 & 85.23$\pm$2.77 & 73.47$\pm$3.83 & 87.44$\pm$2.37 & 74.18 \\
& CCCN & 76.60$\pm$2.02 & 73.48$\pm$2.47 & 70.00$\pm$2.56 & 48.00$\pm$3.63 & 82.28$\pm$1.51 & 50.48$\pm$3.37 & 85.36$\pm$0.73 & 87.44$\pm$1.96 & 74.31$\pm$3.42 & 88.58$\pm$1.73 & 73.65 \\
\midrule
\multirow{4}*{\rotatebox{90}{CCMamba}}
& (graph) & 79.75$\pm$2.40 & \underline{77.42$\pm$4.10} & \underline{73.20$\pm$3.50} & \textbf{52.80$\pm$3.70} & 83.46$\pm$2.40 & \textbf{52.92$\pm$2.85} & \textbf{85.52$\pm$1.10} & \underline{88.12$\pm$0.93} & \underline{76.11$\pm$2.60} & \underline{89.51$\pm$1.87} & \underline{75.88} \\
& (hypergraph) & 78.72$\pm$2.10 & 76.67$\pm$2.30 & 72.40$\pm$2.60 & 46.40$\pm$3.90 & 78.52$\pm$1.80 & 49.53$\pm$2.23 & 83.48$\pm$1.05 & 87.59$\pm$3.16 & 75.03$\pm$4.16 & \textbf{89.53$\pm$2.63} & 73.79 \\
& (simplex) & 76.60$\pm$2.48 & 69.18$\pm$3.15 & 71.20$\pm$2.91 & 42.27$\pm$4.28 & \textbf{85.51$\pm$1.24} & 48.90$\pm$3.53 & 81.12$\pm$1.82 & 70.71$\pm$1.56 & 73.27$\pm$2.63 & 87.61$\pm$2.75 & 70.64 \\
& (rank-cell) & \textbf{85.11$\pm$2.40} & \textbf{78.14$\pm$1.68} & \textbf{74.81$\pm$1.85} & \underline{50.13$\pm$3.02} & \underline{85.49$\pm$1.72} & 50.34$\pm$2.74 & 84.57$\pm$3.10 & \textbf{89.22$\pm$1.50} & \textbf{76.95$\pm$2.90} & 89.29$\pm$1.47 & \textbf{76.41} \\
\bottomrule 
\end{tabular}

\label{tab:1}
\end{table*}

\subsection{Graph Classification and Node Classification}
We evaluate the proposed CCMamba framework on both graph-level and node-level classification tasks, which require leveraging high-order relational structures—such as hypergraphs, simplicial complexes, and cellular complexes
All competing methods are trained under identical hyperparameter configurations to ensure a fair comparison.
We instantiate CCMamba under four topological instantiations, including graph, hypergraph, simplex, and rank-cell.

\subsubsection{\textbf{CCMamba vs. Baseline methods}}
Table~\ref{tab:1} shows that CCMamba (rank-cell) achieves state-of-the-art or competitive performance across multiple graph classification benchmarks. Its core strength lies in the state-space message-passing mechanism, which dynamically filters neighborhood sequences and models long-range dependencies more effectively than traditional aggregation. 
On graph-level tasks, CCMamba(rank-cell) surpasses the strongest cellular baseline CWN by +2.64\% on MUTAG and +3.95\% on PROTEINS, despite utilizing the same cellular complex inputs. Similarly, the graph-based variant, CCMamba (graph), surpasses GAT by +5.86\% on MUTAG. These gains extend to node-level tasks, where CCMamba (rank-cell) reaches 89.22\% on Cora and 76.95\% on Citeseer, exceeding the attention-based CCCN. 
The topological structures also plays a crucial role in performance. CCMamba (rank-cell) achieves the highest mean accuracy (76.41\%) across ten datasets, outperforming the graph-based version (75.88\%) and all hypergraph or simplicial variants. This confirms that higher-order topological relations capture essential information, such as the functional cycles and cliques in MUTAG or complex bond structures in PROTEINS.    Notably, on MUTAG, the performance with +5.36\% gap between CCMamba (rank-cell) and CCMamba (graph) on MUTAG highlights that cliques and cycles provide a more faithful relational mapping for molecular data. 
These results validate CCMamba as a unified, topology-aware message-passing state-space framework capable of exploiting both the efficiency of SSM-based aggregation and the effectiveness of higher-order relational structures.

\begin{table}[htbp]
\centering
\caption{Test accuracy (\%) of GAT, Mamba, MultiGAT, and MultiCCMamba on node-level and graph-level tasks. Best results (\textbf{bold}) and second-best scores (\underline{underlined}).}
\setlength{\tabcolsep}{2pt} 
\begin{tabular}{c |c|c|cc|ccc}
\toprule
 & \multicolumn{2}{c}{Dataset} & GAT & CCMamba & MultiGAT & MultiCCMamba \\
\midrule 
\multirow{12}*{\rotatebox{90}{Node Level Tasks}}
 & \multirow{4}{*}{\rotatebox{90}{\scriptsize Cora}}
 & Graph & 86.71$\pm$2.50 & \underline{88.12$\pm$0.93} & 86.26$\pm$0.79 & \textbf{89.78$\pm$3.21} \\
 & & Hypergraph & 85.05$\pm$1.03 & 75.92$\pm$2.67 & \underline{86.35$\pm$1.37} & \textbf{89.51$\pm$3.22} \\
 & & Simplex& 66.23$\pm$0.71 & \underline{79.17$\pm$1.38} & 69.51$\pm$1.37 & \textbf{79.35$\pm$0.89} \\
 & & Cellular & 86.71$\pm$2.36 & \underline{89.07$\pm$4.77} & 88.47$\pm$3.61 & \textbf{89.52$\pm$2.19} \\
\cmidrule{2-7}
& \multirow{4}{*}{\rotatebox{90}{\scriptsize Citeseer}} & Graph& 73.59$\pm$3.72 & \underline{76.51$\pm$2.90}& 75.87$\pm$1.80 & \textbf{77.43$\pm$1.50}   \\
 & & Hypergraph & 72.45$\pm$1.83& \underline{74.55$\pm$1.55} & 73.71$\pm$1.18 & \textbf{77.43$\pm$3.02}    \\
 & & Simplex& 62.50$\pm$0.94 & 64.33$\pm$1.89 & \underline{65.27$\pm$7.50} & \textbf{69.51$\pm$8.19}    \\
 & & Cellular & 74.97$\pm$4.17& 75.51$\pm$1.93 & \underline{76.35$\pm$2.18} & \textbf{77.31$\pm$4.03 }    \\
\cmidrule{2-7}
& \multirow{4}{*}{\rotatebox{90}{\scriptsize Pubmed}} & Graph& 88.38$\pm$0.28& \underline{89.51$\pm$1.87} & 89.14$\pm$2.51 & \textbf{89.86$\pm$0.41}    \\
 & & Hypergraph & 82.85$\pm$0.40 & \underline{87.89$\pm$3.05} & 87.47$\pm$2.97& \textbf{88.37$\pm$2.51}    \\
 & & Simplex& 62.50$\pm$0.94 & 78.82$\pm$2.95 & \underline{79.39$\pm$3.16} & \textbf{87.36$\pm$4.01}    \\
 & & Cellular & 88.21$\pm$5.27 & 89.29$\pm$1.63 & \textbf{89.46$\pm$3.68} & \underline{89.39$\pm$3.91}    \\
\midrule
\multirow{28}*{\rotatebox{90}{Graph Level Tasks}}
 & \multirow{4}{*}{\rotatebox{90}{\scriptsize MUTAG}} & Graph&72.34$\pm$1.93& \underline{76.60$\pm$2.41} & 73.49$\pm$5.23& \textbf{78.72$\pm$4.86}    \\
& & Hypergraph & 68.72$\pm$12.34 & 72.34$\pm$5.00 & \underline{73.77$\pm$5.63}& \textbf{74.28$\pm$1.05}    \\
 & & Simplex& 66.89$\pm$9.70 & \underline{73.94$\pm$8.20} & 68.08$\pm$10.53 & \textbf{74.47$\pm$8.16}    \\
 & & Cellular & 70.21$\pm$6.03 & \underline{74.47$\pm$4.84} & 71.64$\pm$2.83 & \textbf{76.60$\pm$1.49 }    \\
\cmidrule{2-7}
 & \multirow{4}{*}{\rotatebox{90}{\scriptsize IMDB-M}} & Graph& 48.21$\pm$3.35 & \textbf{52.80$\pm$3.78} & 49.91$\pm$2.57 & \underline{51.20$\pm$3.17}    \\
 & & Hypergraph & 34.19$\pm$1.93 &50.93$\pm$1.47 & 44.93$\pm$2.84 & \textbf{51.98$\pm$0.84}    \\
 & & Simplex& 35.27$\pm$2.01& \underline{37.60$\pm$2.64} & 35.91$\pm$2.56& \textbf{39.62$\pm$3.71 }    \\
 & & Cellular & 48.92$\pm$2.71 & \underline{51.73$\pm$2.83} & 50.13$\pm$2.48& \textbf{53.40$\pm$1.53}    \\
\cmidrule{2-7}
 & \multirow{4}{*}{\rotatebox{90}{\scriptsize PROTEINS}} & Graph& 74.91$\pm$2.47 & \textbf{77.42$\pm$2.33} & 75.25$\pm$1.02 & \underline{76.34$\pm$0.32} \\
 & & Hypergraph & 70.25$\pm$2.08 & \textbf{77.42$\pm$4.88} & 71.24$\pm$2.30 & \underline{76.25$\pm$3.14} \\
 && Simplex& 71.54$\pm$2.20 & \textbf{77.78$\pm$3.67} & 72.71$\pm$1.51 & \underline{75.31$\pm$2.82}  \\
 & & Cellular & 75.13$\pm$0.52 & \underline{76.70$\pm$3.34} & 76.58$\pm$0.93 & \textbf{78.14$\pm$1.05}    \\
\cmidrule{2-7}
 & \multirow{4}{*}{\rotatebox{90}{\scriptsize IMDB-B}} & Graph& 68.40$\pm$3.73& \textbf{73.20$\pm$3.50}& 70.40$\pm$1.51 & \underline{72.43$\pm$1.19} \\
& & Hypergraph & 34.67$\pm$2.08 &\textbf{72.40$\pm$2.60}& 68.40$\pm$2.82 & \underline{72.17$\pm$4.28} \\
 & & Simplex& 45.35$\pm$2.83 & \underline{71.20$\pm$2.91} & 71.04$\pm$2.71 & \textbf{72.80$\pm$3.65} \\
 & & Cellular & 68.62$\pm$3.07 & \textbf{74.81$\pm$1.85} & 70.01$\pm$2.05 & \underline{72.40$\pm$1.12} \\
\cmidrule{2-7}
 & \multirow{4}{*}{\rotatebox{90}{\scriptsize ROMAN-E}} & Graph& \underline{83.25$\pm$1.82} & \textbf{83.46$\pm$2.40} & 79.61$\pm$2.45 &78.43$\pm$1.53 \\
 & & Hypergraph & 71.56$\pm$0.39 & \textbf{78.52$\pm$1.80} & 75.57$\pm$2.61 & \underline{77.25$\pm$2.28} \\
 & & Simplex& 84.70$\pm$2.79 & \textbf{85.51$\pm$1.24} & \underline{84.42$\pm$5.14} & 83.91$\pm$1.82 \\
 & & Cellular & 84.30$\pm$1.85 & 85.49$\pm$1.72 & \underline{86.02$\pm$1.84} & \textbf{87.12$\pm$2.79} \\
\bottomrule 
\end{tabular}

\label{tab:2}
\end{table}

\subsubsection{\textbf{CCMamba vs. Attention mechanism on the higher-order graph learning}}
Table~\ref{tab:2} compares attention-based aggregation (GAT, MultiGAT) with selective state-space aggregation (CCMamba, MultiCCMamba) across node-level and graph-level tasks. To ensure a fair comparison, all models utilize the same backbone architecture and head count ($h=4$), differing only in their message-passing mechanism. 
On node classification tasks,
Mamba-based models outperform their attention-based counterparts across most datasets and structural settings. 
CCMamba achieves notable gains over GAT on simplicial and cellular structures, 
e.g., +12.9\% on Cora (simplex) and +5.0\% on PubMed (hypergraph). MultiCCMamba further amplifies these advantages, attaining the competitive performance with the improvements 
over MultiGAT on Cora (simplex, +9.8\%) and PubMed (simplex, +7.9\%). 
These observed gains demonstrate that the selective state-space mechanism is fundamentally better suited for higher-order representations than standard attention.
On graph-level tasks, similar trends are observed. CCMamba outperforms GAT by substantial margins, particularly on IMDB-B (hypergraph, +37.7\%) and IMDB-M (hypergraph, +16.7\%).
In these cases, attention-based models often degrade whereas CCMamba remains robust, such as GAT collapses to 34.67\% on the IMDB-B hypergraph. MultiCCMamba also sets the benchmark on ROMAN-E (cellular, 87.12\%), outperforming MultiGAT by +1.1\%. On PROTEINS, CCMamba achieves peak accuracy across all four topological structures. These results confirm that SSM-based aggregation provides a consistent and scalable advantage over attention for structured data.

\subsubsection{Expressive Power of CCMamba}
Table~\ref{tab:3} evaluates the expressive power of CCMamba(GIN) by comparing it against isomorphism-based neural networks: GIN (graph), UniGIN (hypergraph), 
and CWN (cellular complex). All models utilize a GIN-style backbone to isolate the impact of topological structure from the aggregation mechanism. CCMamba(GIN) consistently matches or surpasses all baselines across both graph-level and node-level tasks. Compared to the standard GIN, the most significant improvements occur on structurally complex datasets, such as ROMAN-E (+9.01\%), and IMDB-M (+10.50\%).
Since both models share an identical backbone, these improvements demonstrate that rank-aware SSM sequences capture structural patterns strictly beyond the capacity of 1-WL-equivalent pairwise message passing. Furthermore, CCMamba (GIN) achieves substantial leads over UniGIN on IMDB-M (+56.84\%) and ROMAN-E (+18.85\%), highlighting that symmetric hyperedge pooling without rank-aware directionality fails to resolve multi-rank structural patterns. Compared to CWN, the notable improvements on ROMAN-E (+5.53\%) and Citeseer (+4.90\%) suggest that local boundary-operator convolutions are less effective than SSMs at modeling long-range directional dependencies. 
This empirical evidence confirms that CCMamba activates expressiveness beyond both pairwise and hyperedge-level isomorphism tests. These results are consistent with Theorem~1, which establishes that CCMamba is upper-bounded by the 1-CCWL test.

\begin{table}[htbp]
\caption{Testing accuracies (\%) of GIN, UniGIN, CWN and CCMamba(GIN) on graph benchmarks (mean$\pm$std). Best results (\textbf{bold}) and second-best scores (\underline{underlined}).}
\setlength{\tabcolsep}{3pt}
\centering
\begin{tabular}{lccccc}
\toprule 
\textbf{Dataset } & \textbf{GIN} & \textbf{UniGIN} & \textbf{CWN} & \textbf{CCMamba(GIN)} \\
\midrule
MUTAG & 78.72$\pm$2.61 & \textbf{79.23$\pm$4.85} & 82.47$\pm$1.34 & \underline{78.88$\pm$3.75} \\
PROTEINS & \underline{76.70$\pm$1.87}& 71.26$\pm$2.09 & 74.19$\pm$2.27 &\textbf{77.42$\pm$2.71} \\
IMDB-B & 71.20$\pm$1.86 &68.48$\pm$2.17 & \underline{72.20$\pm$2.17} &\textbf{72.40$\pm$2.31} \\
IMDB-M & 46.40$\pm$4.07 &32.69$\pm$1.06 & \underline{49.07$\pm$3.42} & \textbf{51.27$\pm$3.03} \\
ROMAN-E& 79.92$\pm$2.20 & 73.29$\pm$0.28 & \underline{82.55$\pm$1.27} & \textbf{87.12$\pm$0.99} \\
AMAZON-R & 49.22$\pm$3.58 & 48.75$\pm$0.43 & \textbf{50.99$\pm$3.14} & \underline{50.08$\pm$1.28} \\
Minesweeper & 83.60$\pm$0.37 & 80.36$\pm$0.79 & \underline{84.00$\pm$0.73} & \textbf{84.88$\pm$0.93} \\
\midrule
Cora & \underline{87.59$\pm$1.92} & 84.64$\pm$0.78 & 85.23$\pm$2.77 & \textbf{88.53$\pm$1.39} \\
Citeseer & 73.11$\pm$4.08 &72.53$\pm$1.09& \underline{73.47$\pm$3.83} &\textbf{77.07$\pm$3.26} \\
PubMed & \underline{89.51$\pm$0.94} &83.08$\pm$0.41 & 87.44$\pm$2.37 & \textbf{89.61$\pm$1.48} \\
\bottomrule 
\end{tabular}

\label{tab:3}
\end{table}

\subsection{Parameter Sensitivity}

\begin{table}[htbp]
\centering
\caption{Testing accuracies(\%) of multiple CCMamba variants with different layer depths (2,4,8,16,32,64) in node classification task. Best results (\textbf{bold}) and second-best scores (\underline{underlined}).}
\setlength{\tabcolsep}{3pt} 
\begin{tabular}{clcccccc}
\toprule 
\multirow{2}{*}{Datasets} & \multirow{2}{*}{Models} & \multicolumn{5}{c}{Layers} \\
\cmidrule{3-8}
 & & 2 & 4 & 8 & 16 & 32 & 64  \\
\midrule
\multirow{6}{*}{Cora} & GCN & 87.00 & 85.23 & 82.27 & 79.24 & 74.43 & 73.26 \\
 & HyperGCN & 84.34 & 79.62 & 12.70 & 12.38 & 12.85 & 13.29 \\
 & SCCN & 71.34 & 68.83 & 54.51 & 47.27 & 41.65 & 55.69 \\
 & CCCN & 88.18 & 86.56 & 84.34 & 82.57 & 77.84& 74.30 \\
 \cmidrule{2-8}
 & CCMamba(graph) & 87.44 & \textbf{88.77} & \textbf{87.89} & \textbf{88.04} & \textbf{87.74} & \textbf{78.07} \\
 & CCMamba(hypergraph) &\underline{88.92} & \underline{88.48} & 85.52& \underline{76.96} & 76.51 & \underline{77.40}\\
 & CCMamba(simplex) &86.51 & 84.50&75.48 & 73.26 & 73.56 & 74.13 \\
 & CCMamba(rank-cell) & \textbf{89.07} & 88.33 & \underline{86.07} & 76.66 & \underline{77.40} & 76.07 \\
\midrule
\multirow{6}{*}{Citeseer} & GCN & 75.75 & 73.59 & 70.47 & 69.75 & 69.15 & 67.83 \\
 & HyperGCN & 72.75 & 70.71 & 8.40 & 7.44 & 8.40 & 7.68 \\
 & SCCN & 66.39 & 64.95 & 62.91 & 45.86 & 39.86 &54.26 \\
 & CCCN & 76.23 & 72.75 & 70.11 & 68.79 & 68.91 & 68.55 \\
 \cmidrule{2-8}
 & CCMamba(graph) & 76.11 & \textbf{77.43} & \textbf{76.47} & \textbf{74.43} & \underline{74.31} & 72.63 \\
 & CCMamba(hypergraph) &\textbf{76.95} & \underline{75.99} &\underline{76.23}& \underline{73.59} & \textbf{74.79} & \textbf{74.07} \\
 & CCMamba(simplex) & 67.23 &69.27 & 72.15 & 70.83 & 71.43 & 68.79 \\
 & CCMamba(rank-cell) & \underline{76.51} & 74.91 & 73.59 & 73.47 & 72.87 & \underline{73.62} \\
\midrule
\multirow{6}{*}{Pubmed} & GCN & 89.55 & 88.44 & 88.99 & 84.86 & 81.44 & 75.80 \\
 & HyperGCN & 82.82 & 78.05 & 21.30 & 19.78 & 19.78 & 19.53 \\
 & SCCN & 86.67 & 86.04 & 66.80 & 57.78 & 58.26 & 61.08 \\
 & CCCN & 88.58 & 87.36 & 86.21 & 85.44 & 84.72 & 85.14 \\
 \cmidrule{2-8}
 & CCMamba(graph) & 89.51 & \underline{89.68} & \underline{89.23} & \underline{88.95} & \underline{88.71} & 84.47 \\
 & CCMamba(hypergraph) &\textbf{89.94} & \textbf{89.78} & \textbf{90.13} & \textbf{89.15} & \textbf{88.94} & \underline{87.48} \\
 & CCMamba(simplex) & 87.30 & 89.01 & 87.34 & 87.12 & 83.08& 83.57 \\
 & CCMamba(rank-cell) & \underline{89.83} & 89.41 & {88.76} & {88.12} & {87.65} & \textbf{87.51} \\
\bottomrule 
\end{tabular}
\label{tabv}
\end{table}

\subsubsection{Analysis of Deep-layered CCMamba }
Table~\ref{tabv} evaluates CCMamba's robustness against over-smoothing as depth increases from 2 to 64 layers. Conventional GCNs and higher-order convolutional models (HyperGCN, SCCN, CCCN) suffer significant performance degradation with increased depth. For example, on Cora, HyperGCN collapses from 84.34\% to 12.70\% at 8 layers ($-84.94\%$), while SCCN drops to 41.65\% at 32 layers
($-$41.60\% ).  GCN also exhibits a consistent decline, falling to 73.26\% at 64 layers ($-15.79\%$). In contrast, CCMamba remains remarkably stable. On Cora, CCMamba (graph) maintains consistent accuracy from 2 to 32 layers (87.44\% vs. 87.74\%), showing only marginal decay even at extreme depths. This stability stems from the selective state-space mechanism, which replaces passive local averaging with input-dependent transitions. By adaptively gating neighborhood information, CCMamba preserves discriminative features across deep architectures, effectively mitigating the over-smoothing problem inherent in topological message passing.

\begin{table}[htbp]
\centering
\caption{Test accuracy (\%) of Multi-heads CCMamba with different heads number (e.g., 2,4,8,16) on node- and graph-level tasks. Best results (\textbf{bold}) and second-best scores (\underline{underlined}).}
\setlength{\tabcolsep}{2pt} 
\begin{tabular}{c l cccc}
\toprule
\multirow{2}{*}{Datasets} & \multirow{2}{*}{Models} & \multicolumn{4}{c}{ Multi-Heads Number} \\
\cmidrule{3-6}
 & & 2 & 4 & 8 &
 16  \\
\midrule
\multirow{10}*{\rotatebox{90}{CCMamba-graph}} & MUTAG & \textbf{89.15$\pm$0.56} & 76.60$\pm$4.05 & \underline{80.85$\pm$7.69} & 74.47$\pm$3.58 \\
  & PROTEINS & 75.99$\pm$1.53 & \underline{77.06$\pm$2.65} & 77.06$\pm$1.93 & \textbf{79.21$\pm$2.60} \\
  & IMDB-B & \textbf{74.00$\pm$4.69} & 73.20$\pm$2.94 & \underline{74.00$\pm$3.26} & 73.60$\pm$2.69 \\
  & IMDB-M & 50.93$\pm$2.16 & \underline{53.87$\pm$3.36} & \textbf{54.93$\pm$3.87} & 50.67$\pm$3.12 \\
  & ROMAN-E & \textbf{85.14$\pm$0.63} & \underline{84.17$\pm$0.20} & 84.01$\pm$0.23 & 84.15$\pm$0.26 \\
  & AMAZON-R & \textbf{53.98$\pm$0.76} & 53.45$\pm$0.87 & 52.67$\pm$0.81 & \underline{53.70$\pm$0.84} \\
  & Minesweeper & 85.88$\pm$0.56 & \textbf{86.60$\pm$0.93} & 85.92$\pm$0.67 & 85.88$\pm$0.46 \\
  & Cora & \textbf{85.97$\pm$1.16} & 85.52$\pm$1.03 & 84.93$\pm$1.15 & \underline{85.97$\pm$1.15} \\
  & Citeseer & 74.31$\pm$1.26 & 74.55$\pm$0.93 & \underline{74.79$\pm$1.53} & \textbf{75.51$\pm$1.84} \\
  & PubMed & 89.47$\pm$0.28 & \underline{89.55$\pm$0.21} & 89.47$\pm$0.07 & \textbf{89.57$\pm$0.31} \\
\midrule
\multirow{10}*{\rotatebox{90}{CCMamba-hypergraph}} & MUTAG & 76.60$\pm$3.84 & 76.60$\pm$4.85 & \underline{78.72$\pm$9.30} & \textbf{78.72$\pm$4.56} \\
  & PROTEINS & 74.55$\pm$2.50 & 74.55$\pm$2.50 & \underline{75.27$\pm$2.06} & \textbf{77.78$\pm$1.63} \\
  & IMDB-B & \underline{73.20$\pm$1.84} & 72.40$\pm$2.93 & 71.20$\pm$2.38 & \textbf{75.20$\pm$4.18} \\
  & IMDB-M & 52.53$\pm$2.40 & 52.00$\pm$1.63 & \textbf{53.87$\pm$2.91} & \underline{53.33$\pm$3.94} \\
  & ROMAN-E & \textbf{82.05$\pm$0.32} & \underline{80.96$\pm$0.42} & 79.69$\pm$0.06 & 78.77$\pm$0.26 \\
  & AMAZON-R & \underline{51.53$\pm$0.67} & 50.94$\pm$0.67 & 51.36$\pm$0.62 & \textbf{52.02$\pm$1.03} \\
  & Minesweeper & 80.84$\pm$0.86 & 83.20$\pm$1.84 & \textbf{83.96$\pm$0.59} & \underline{83.68$\pm$1.92} \\
  & Cora & 87.44$\pm$0.69 & 87.89$\pm$0.83 & \underline{88.63$\pm$1.01} & \textbf{88.77$\pm$1.32} \\
  & Citeseer & \textbf{77.31$\pm$1.82} & 76.71$\pm$1.20 & \underline{77.31$\pm$1.25} & 77.19$\pm$1.86 \\
  & PubMed & 89.57$\pm$0.40 & 89.59$\pm$0.42 & \textbf{89.88$\pm$0.53} & \underline{89.72$\pm$0.48} \\
\midrule
\multirow{10}*{\rotatebox{90}{CCMamba-simplex}} & MUTAG & \textbf{85.11$\pm$9.74} & \underline{80.85$\pm$11.39} & 76.60$\pm$4.95 & 72.34$\pm$5.06 \\
  & PROTEINS & \textbf{73.12$\pm$3.60} & \underline{72.76$\pm$1.49} & 72.76$\pm$1.09 & 71.68$\pm$1.53 \\
  & IMDB-B & \textbf{75.20$\pm$3.69} & \underline{71.20$\pm$2.65} & 68.80$\pm$0.57 & 69.25$\pm$0.53 \\
  & IMDB-M & 44.53$\pm$5.70 & 35.20$\pm$0.89 & 42.67$\pm$5.09 & \textbf{48.27$\pm$11.50} \\
  & ROMAN-E & \textbf{82.46$\pm$0.91} & \underline{81.26$\pm$0.59} & 80.50$\pm$1.32 & 80.74$\pm$1.54 \\
  & AMAZON-R & \textbf{48.57$\pm$0.86} & \underline{47.84$\pm$4.30} & 46.96$\pm$2.72 & 47.27$\pm$1.96 \\
  & Minesweeper & 82.32$\pm$1.49 & \textbf{82.76$\pm$2.30} & 82.20$\pm$1.37 & \underline{82.61$\pm$1.96} \\
  & Cora & \textbf{67.65$\pm$4.93} & 40.62$\pm$3.17 & 44.31$\pm$6.25 & \underline{54.95$\pm$10.37} \\
  & Citeseer & \textbf{65.55$\pm$3.70} & \underline{51.62$\pm$9.48} & 50.66$\pm$7.84 & 45.02$\pm$2.63 \\
  & PubMed & \underline{86.88$\pm$0.47} & \textbf{87.63$\pm$0.42} & 85.96$\pm$3.30 & 85.86$\pm$17.93 \\
\midrule
\multirow{10}*{\rotatebox{90}{CCMamba-rank-cell}} & MUTAG & \underline{80.85$\pm$5.64} & 76.60$\pm$4.58 & 80.85$\pm$4.16 & \textbf{82.98$\pm$3.85} \\
  & PROTEINS & \textbf{78.14$\pm$2.39} & 77.42$\pm$2.65 & \underline{77.78$\pm$3.15} & 77.78$\pm$2.69 \\
  & IMDB-B & 72.00$\pm$2.15 & \underline{72.80$\pm$2.31} & 72.80$\pm$1.36 & \textbf{72.80$\pm$1.08} \\
  & IMDB-M & 51.20$\pm$2.68 & \textbf{53.87$\pm$2.96} & 49.33$\pm$1.72 & \underline{51.47$\pm$2.93} \\
  & ROMAN-E & \textbf{85.70$\pm$0.69} & 85.42$\pm$0.21 & \underline{85.51$\pm$0.40} & 84.86$\pm$0.27 \\
  & AMAZON-R & 48.88$\pm$1.15 & \textbf{49.98$\pm$0.83} & \underline{49.11$\pm$0.65} & 48.83$\pm$0.69 \\
  & Minesweeper & 83.32$\pm$0.67 & 83.92$\pm$0.65 & \textbf{83.96$\pm$0.65} & \underline{83.96$\pm$0.57} \\
  & Cora & \textbf{89.51$\pm$1.63} & 88.04$\pm$0.83 & \underline{88.48$\pm$0.62} & 88.18$\pm$0.92 \\
  & Citeseer & 76.71$\pm$1.23 & \underline{77.19$\pm$1.58} & 77.07$\pm$1.82 & \textbf{77.43$\pm$1.78} \\
  & PubMed & 88.82$\pm$0.49 & 88.86$\pm$0.45 & \underline{88.97$\pm$0.53} & \textbf{89.15$\pm$0.56} \\
\toprule
\end{tabular}

\label{tables3}
\end{table}

\subsubsection{Sensitivity of Multi-heads Numbers} Table 
\ref{tables3} evaluates CCMamba with $\{2, 4, 8, 16\}$ head configurations four topological domains.
For the graph topology, fewer heads yield the most competitive performance. On MUTAG (graph), accuracy peaks at 89.15\% (heads =2) and degrades to 74.47\% (heads=16), since excessive head partitioning fragments the per-head representational capacity below the threshold required to preserve localized structural features. Meanwhile, CCMamba (simplex) exhibits even sharper sensitivity: on Cora, accuracy collapses from 
67.65\% (heads =2) to 40.62\% (heads =4). In contrast, CCMamba(hypergraph) benefits from larger head numbers, with Cora accuracy rising from 87.44\% (heads=2) to 88.77\% (heads=16).
CCMamba(rank-cell) remains the most stable performance across all head numbers, maintaining stable performance across all 
configurations (e.g., PubMed: 88.82\% to 89.15\% from heads number 2 to 16), as the rank functions regularize information across topological dimensions independently of head.

\begin{figure*}[htbp]
 \centering
\includegraphics[width=1\linewidth]{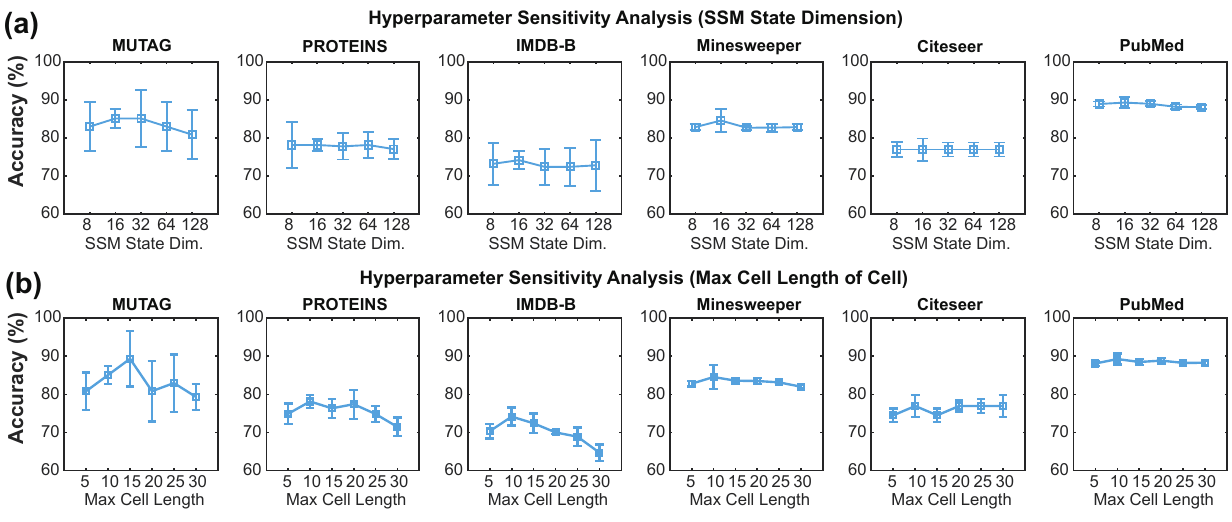}
 \caption{Parameters sensitivity analysis. (a) The SSM State Dimension ($d_state$) and  maximum cell length of cell construction on different datasets}
 \label{fig04}
\end{figure*}

\subsubsection{Sensitivity of SSM State Dimension} 

Figure \ref{fig03}(a) shows the impact of the hidden state dimension $d_{\text{state}}$ of the SSM on CCMamba performance. This parameter controls the memory capacity of the selective state-space model and its ability to preserve long-range topological dependencies during order-aware sequence propagation. 
On most datasets, performance consistently peaks at $d_{\text{state}}=16$ and decreases with increasing dimensionality. On the MUTAG dataset, accuracy reaches 85.11\% at $d_{\text{state}}=16$, subsequently dropping to 80.85\% (a relative decrease of 5.01\%) at $d_{\text{state}}=128$. The IMDB-B dataset similarly peaks at $d_{\text{state}}=16$ at 74.16\%, decreasing to 72.40\% with increasing dimensionality. Minesweeper exhibits the most acute sensitivity, peaking at 84.57\% with $d_{\text{state}}=16$, but declining to 82.85\%. 
In contrast, Citeseer and PubMed remain stable across all state dimensions. When $d_{\text{state}}$ is too small (e.g., 8), SSM lacks sufficient memory capacity to preserve discriminative topological signals in long sequences. Conversely, excessively large state dimensions (e.g., 64) over-parameterize state transition dynamics, making them overly complex relative to the complex structure of the input, thus increasing optimization difficulty and the risk of overfitting on small or medium-sized datasets (e.g., MUTAG). The performance of node-level citation datasets further confirms that their homogeneous local topology requires only limited SSM memory to propagate sufficient structural context.

\subsubsection{Sensitivity to Maximum Cell Size}
\begin{figure*}[htbp]
 \centering
\includegraphics[width=0.95\linewidth]{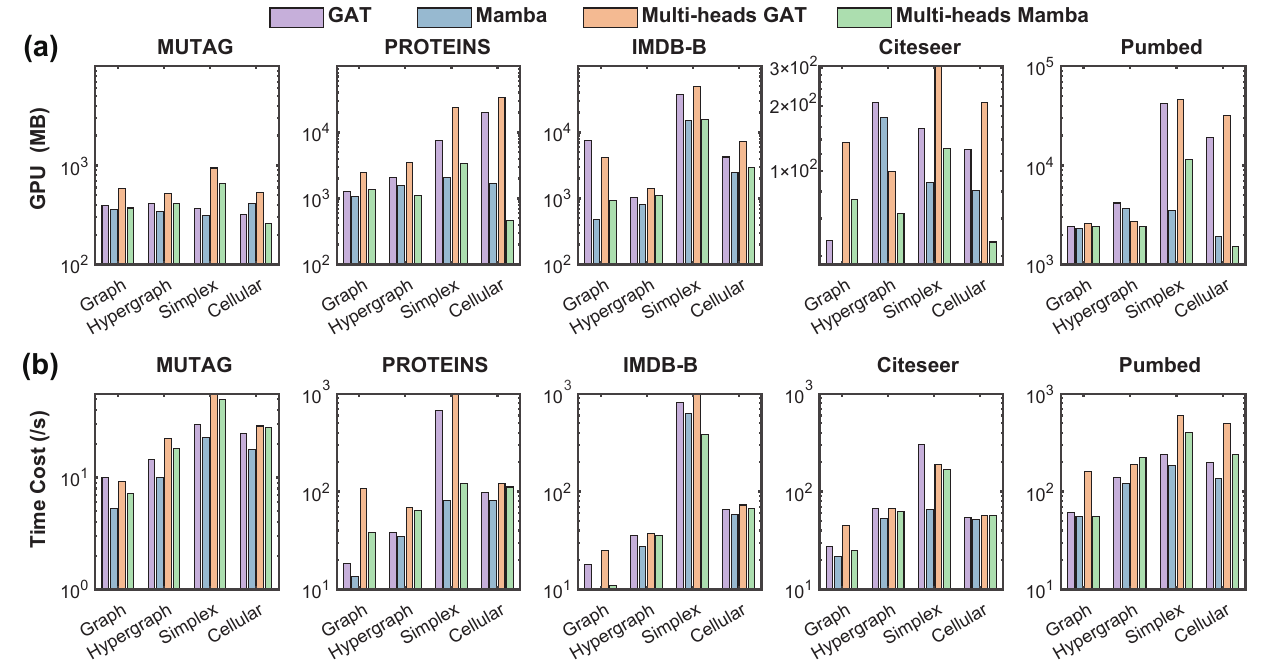}
 \caption{Comparison of GPU Memory (MB) and Time Cost(/s) on different datasets.}
 \label{fig03}
\end{figure*}

Figure \ref{fig03}(b) illustrates the impact of the maximum cell length ($S$) on CCMamba performance. This parameter controls the maximum number of nodes allowed in each higher-order cell during the graph lift process to higher-order structures. When $S$ is too small, the lift operation produces over-fragmented cells that cannot encode multi-node higher-order interactions, thus limiting the structural expressiveness of the complex. Conversely, overly large cells aggregate redundant nodes into a single higher-order cell, blurring the structure and introducing noise into SSM aggregation. On the MUTAG dataset, accuracy increases from 80.85\% at $S$=5 to a peak of 89.36\% (+10.52\%) at $S$=15, then decreases to 79.35\% (-11.28\%) at $S$=30. Similarly, accuracy on PROTEINS, IMDB-B, and Minesweeper datasets monotonically decreases as cell size gradually peaks, while node-level datasets Citeseer and PubMed remain relatively stable across all sizes. The maximum cell length further reflects the differences in the topology of the underlying graphs. Molecular graphs (such as MUTAG) benefit from larger cells that can capture cyclic structures; while citation networks stable at smaller cell sizes for their denser and more uniform local connections.

\begin{table}[htbp]
\centering
\caption{Performance sensitivity analysis of CCMamba across different input dimensions. Best results (\textbf{bold}) and second-best scores (\underline{underlined}).}
\label{tablevii}
\begin{tabular}{lccc} 
\toprule 
Input dim of rank & {[}0{]} & {[}0,1{]} & {[}0,1,2{]} \\
\midrule
MUTAG & \underline{79.75±2.40} & \textbf{85.11±2.40} & —  \\
PROTEINS & \underline{77.42±4.10} & \textbf{78.14±1.68} & —  \\
IMDB-B & 73.20±3.50 & \underline{74.16±2.35} &\textbf{ 74.81±1.85} \\
IMDB-M & \textbf{52.80±3.70} & \underline{50.13±3.02} & —  \\
ROMAN-E & 83.46±2.40 & \underline{85.09±0.40} & \textbf{85.67±2.94} \\
AMAZON-R & \textbf{52.92±2.85} & \underline{50.34±2.74} & —  \\
Minesweeper & \textbf{85.52±1.10} & 82.55±1.15 & \underline{84.57±3.10} \\
Cora & \underline{88.12±0.93} & 87.33±1.79 & \textbf{89.22±1.50} \\
Citeseer & \underline{76.11±2.60} & 74.50±1.75 & \textbf{76.95±2.90} \\
PubMed & \textbf{89.51±1.87} & 88.04±0.51 & \underline{89.29±1.47} \\
\bottomrule
\end{tabular}
\end{table}

\subsubsection{Analysis of Input Rank Dimension}
Table \ref{tablevii} analyzes the performance sensitivity of CCMamba (rank-cell) as the input rank dimension increases from rank-0 only $[0]$, to rank-0 and rank-1 $[0,1]$, and to the full rank $[0,1,2]$, which further evaluates whether incorporating higher-order cells.  
On most datasets, CCMamba(rank-cell) incorporates higher-order topological dimensions that improves prediction accuracy. On MUTAG, extending from $[0]$ to $[0,1]$ brings a relative gain of +6.71\%, and on ROMAN-E the full rank $[0,1,2]$ achieves the best accuracy (85.67\%), outperforming the  rank-0 baseline by +2.65\%. For node classification, 
$[0,1,2]$ attains the highest accuracy on Cora (89.22\% vs.\ 88.12\%), Citeseer (76.95\% vs.\ 76.11\%), which confirms that face-level incidence provide complementary structural information beyond pairwise edges. This shows that CCMamba can effectively utilize the incident relationship in the combinatorial complex to capture higher-order structural features such as cycles and cliques. These results validate 
that the benefit of CCMamba's rank-aware design allows to exploit multi-rank incidence structures selectively where they carry discriminative topological information.

\subsection{\textbf{Efficiency Analysis}.} 

\textbf{SSM vs. Attention.}  Figure~\ref{fig03} compares the GPU memory consumption (MB) and time cost (/s) of GAT, 
CCMamba, Multi-heads GAT, and Multi-heads CCMamba across five datasets under four topological structures, which evaluates the practical scalability of SSM-based aggregation against attention-based counterparts. Figure~\ref{fig03}(a) shows CCMamba effectively reduces GPU memory relative to GAT across most datasets and topological settings. On higher-order representations where attention complexity peaks,  , CCMamba achieves dramatic memory savings 72.4\% on PROTEINS (simplex), and 60.4\% on IMDB-B (simplex). This efficiency gap widens with multi-head architectures. On PROTEINS, Multi-head GAT consumes 23,518 MB, while Multi-head CCMamba requires only 3,342 MB (85.8\% reduction). Figure~\ref{fig03} displays training speeds show similar gains, such as CCMamba reduces time by 87.9\% on PROTEINS and 22.3\% on IMDB-B. 
Unlike attention, multi-head SSMs partition the state space across heads without replicating quadratic attention maps, maintaining sub-linear memory growth. These results establish CCMamba and Multi-head CCMamba as scalable alternatives for high-order topological message passing, enabling efficient learning on large-scale complexes structured data.

\section{Conclusion}
This paper proposes Combinatorial Complex Mamba (CCMamba), which reformulates message passing on combinatorial complexes from a state-space modeling perspective. In contrast to existing TDL methods that rely on attention-based aggregation and suffer from quadratic complexity, CCMamba provides an efficient and expressive alternative. By enabling rank-aware and long-range information propagation with linear complexity, CCMamba generalizes across graphs, hypergraphs, simplicial complexes, and cellular complexes, while achieving expressive power characterized by the 1-CCWL test. Moreover, CCMamba displays strong robustness to depth, effectively mitigating over-smoothing problem. Overall, our results on real-data benchmarks indicate that structured state-space models constitute a scalable and principled foundation for higher-order representation learning beyond graphs.

\bibliographystyle{IEEEtran} 
\bibliography{IEEEtran}

\end{document}